\documentclass[11pt]{article}

\usepackage[final]{acl}

\usepackage[utf8]{inputenc}   
\usepackage[T1]{fontenc}    	
\usepackage{url}              		 
\usepackage{booktabs}       	
\usepackage{amsfonts}       	
\usepackage{nicefrac}       	  
\usepackage{microtype}      	
\usepackage{subcaption} 

\usepackage{graphicx}
\usepackage{color}
\usepackage{xcolor}
\usepackage{rotating}
\usepackage{tabularx}
\usepackage{pdflscape}
\usepackage{amsmath,amsthm,amssymb}
\usepackage{algorithm, algorithmic}
\usepackage{bbm,dsfont}
\usepackage{caption}
\usepackage{wrapfig}
\usepackage{cancel}
\usepackage{enumerate,cases}
\usepackage{thmtools,thm-restate}
\usepackage{mathtools}

\usepackage{multirow}
\usepackage{makecell}
\usepackage{setspace}
\usepackage{pifont}
\usepackage{array}
\usepackage{enumitem}

\allowdisplaybreaks

\usepackage{hyperref}		 
\hypersetup{
}
\usepackage[capitalize]{cleveref}
\crefformat{figure}{Fig.~#2#1#3}       
\crefformat{table}{Tab.~#2#1#3}        
\crefformat{equation}{Eq.~(#2#1#3)}    
\crefformat{section}{Sec.~#2#1#3}      
\crefformat{appendix}{App.~#2#1#3}     
\crefformat{algorithm}{Alg.~#2#1#3}    

\usepackage{silence}
\usepackage{times}
\usepackage{latexsym}
\usepackage{inconsolata}
\usepackage{titletoc}

\newcommand{\para}[1]{\noindent\textbf{#1}~}

\renewcommand{\phi}{\varphi}
\renewcommand{\epsilon}{\varepsilon}

\newcommand{\step}[1]{{\large \textcircled{\small #1}}}

\newcommand{\al}[1]{ \begin{align} #1  \end{align}}
\newcommand{\eq}[1]{ \begin{equation} #1  \end{equation}}
\newcommand{\als}[1]{ \begin{align*} #1  \end{align*}}
\newcommand{\eqs}[1]{ \begin{equation*} #1  \end{equation*}}

\newcommand{\Lp}{\left(}
\newcommand{\Rp}{\right)}
\newcommand{\Lb}{\left[}
\newcommand{\Rb}{\right]}

\newcommand{\el}{\end{flushleft}}
\newcommand{\bl}{\begin{flushleft}}

\newcommand{\squishlisttwo}{
 \begin{list}{$\bullet$}
  { \setlength{\itemsep}{1pt}
     \setlength{\parsep}{0pt}
    \setlength{\topsep}{0pt}
    \setlength{\partopsep}{0pt}
    \setlength{\leftmargin}{1em}
    \setlength{\labelwidth}{1.5em}
    \setlength{\labelsep}{0.5em} } 
}
\newcommand{\squishend}{\end{list}}

\newcommand{\cB}{\mathcal{B}}

\newcommand{\cD}{\mathcal{D}}

\newcommand{\cG}{\mathcal{G}}

\newcommand{\cL}{\mathcal{L}}
\newcommand{\cM}{\mathcal{M}}
\newcommand{\cN}{\mathcal{N}}

\newcommand{\cP}{\mathcal{P}}

\newcommand{\cR}{\mathcal{R}}

\theoremstyle{plain}

\title{
    Scaling Near-Optimal SFT--RL Annotation Budget Allocation \\ from Small to Large LLMs
}

\author{
    \textbf{Jingtan Wang\textsuperscript{1,3}},
    \textbf{Arun Verma\textsuperscript{2,$\star$}},
    \textbf{Xiaoqiang Lin\textsuperscript{1}},
    \textbf{Zhengyuan Liu\textsuperscript{3}}, \\
    \textbf{Nancy F. Chen\textsuperscript{3}},
    \textbf{Daniela Rus\textsuperscript{4,2}},
    \textbf{Bryan Kian Hsiang Low\textsuperscript{1,2}}
    \\
    {\small
        \textsuperscript{1}Department of Computer Science, National University of Singapore, Republic of Singapore
    }
    ~\\
    {\small
        \textsuperscript{2}Singapore-MIT Alliance for Research and Technology Centre, Republic of Singapore
    }
    ~\\
    {\small
        \textsuperscript{3}Agency for Science, Technology, and Research (A*STAR), Republic of Singapore 
    }
    ~\\
    {\small
        \textsuperscript{4}CSAIL, Massachusetts Institute of Technology, USA 
    }
}

\begin{document}

    \maketitle
    \makeatletter
    \def\blankfootnote{\xdef\@thefnmark{}\@footnotetext}
    \makeatother

    \begin{abstract}

How to divide a fixed annotation budget between supervised fine-tuning (SFT) and reinforcement learning (RL) during LLM post-training remains an open problem. Existing work characterizes only broad trends (e.g., SFT dominates in low-data regimes), lacks a principled allocation framework, and does not examine whether the optimal ratio transfers across model sizes. We frame this problem in terms of near-optimality: rather than seeking a single optimal SFT--RL ratio, we characterize the near-optimal region, the set of allocations within a specified tolerance of peak performance. Empirically, this region is wide even for small tolerances ($2$--$10\%$),  widens with model scale, and transfers reliably from small proxy models to large target models. This yields a practical strategy: small proxy-model experiments suffice to identify a transferable near-optimal region, eliminating the need for exhaustive large-scale search. Our results hold consistently across tasks, model families, and both preference-based off-policy and reward-supervision on-policy RL methods. We further analyze how the asymmetry in annotation costs between SFT and RL data shifts the near-optimal region.

        \blankfootnote{\textbf{$^\star$}Corresponding author: arun.verma@smart.mit.edu.}
    \end{abstract}


    \section{Introduction}
    \label{sec:introduction}

Post-training has become an essential stage for equipping pretrained models with targeted downstream capabilities~\citep{ivison2023camels, lamberttulu3}. 
Modern LLM post-training pipelines are multi-stage, typically consisting of supervised fine-tuning (SFT) followed by reinforcement learning (RL) or preference-based optimization~\citep{guo2025deepseek}. 
This structure raises a natural question: \emph{given a fixed annotation budget for training samples, how should it be divided between SFT and RL?} 
Allocating too few samples to SFT leaves a weak initial policy that RL cannot effectively improve; over-investing in SFT reduces the budget available for preference and reward supervision data necessary for policy improvement.
The right balance depends on model scale, task, and post-training method, yet determining it without exhaustive experimentation remains an open problem.

The most relevant work, \citet{raghavendra2025balancing}, characterizes only a broad trend (SFT dominates in the low-data regime while preference optimization becomes more favorable at scale), and neither identifies the optimal allocation without grid search nor establishes whether it transfers across model scales. 
Other related work on data mixture optimization addresses only single-stage settings: domain mixtures during pretraining~\citep{xie2023doremi,fan2024doge} or via scaling-law extrapolation~\citep{liu2025regmix, ye2025data}; instruction-mixture composition during fine-tuning~\citep{ivison2023camels, lamberttulu3, shi2025damo, corrado-etal-2025-automixalign}; and the pretraining--fine-tuning forgetting trade-off~\citep{bethune2025scaling}. 
None of these addresses multi-stage SFT--RL budget allocation or how to optimize it at scale.

SFT--RL budget allocation problem presents two fundamental challenges. \step{1}~Post-training is sensitive to fine-tuning method and task~\citep{zhang2024when}, metric construction~\citep{schaeffer2025why}, reward overoptimization~\cite{gao2023scaling}, and inverse scaling~\cite{mckenzieinverse}; moreover, post-training algorithm rankings can flip across scale~\citep{li2026post}.
Although certain aggregate behaviors remain predictable~\citep{gadre2025language,ruan2024observational}, the optimal \emph{SFT--RL ratio may not transfer reliably across model scales} (\cref{sec: scaling-law}). 
\step{2}~Any adjustment to the SFT--RL allocation typically requires retraining from scratch or rerunning subsequent stages, making \emph{exhaustive grid search at full model scale prohibitively expensive}.

To address these challenges, we frame the problem in terms of \emph{near-optimality} rather than exact optimality. 
Instead of seeking a single optimal ratio, we introduce the \emph{near-optimal region}: the set of allocations whose performance lies within a fixed tolerance of the best observed performance (\cref{sec: epsilon-region}). 
Our experimental results show that this region is wide even for small tolerances ($2$--$10\%$ of peak performance), grows consistently wider with model scale, and transfers reliably from small proxy models to large target models. 
These results yield a practical strategy: small proxy-model experiments suffice to identify a transferable near-optimal region, eliminating the need for exhaustive large-scale search (\cref{sec: scaling-law}).

\vspace{1mm}\noindent 
We summarize our key contributions as follows:
\vspace{-1mm}
\begin{itemize}[leftmargin=1em]
    \setlength\itemsep{0.0em}
    \item \textbf{Near-optimal region analysis.} We show that even a small tolerance (e.g., a few percent of the optimum) yields a wide near-optimal region rather than a single sharp optimum across tasks and model families (\cref{sec: epsilon-region}).  
    
    \item \textbf{Scale-dependent region expansion.} At a fixed tolerance, the near-optimal region generally widens with model scale; regions identified on small proxy models transfer reliably to larger targets, enabling efficient region discovery without exhaustive large-scale search (\cref{sec: scaling-law}).

    \item \textbf{Generality across settings.} 
    The near-optimal region widens consistently across tasks, model families, and both off-policy and on-policy RL methods. We further show how annotation cost asymmetry between SFT and RL data shifts the optimal allocation (\cref{sec: money}): the region widens as SFT becomes relatively more expensive.
\end{itemize}

    \section{Preliminaries}
    \label{sec:problem}

\subsection{Problem Formulation}
\label{sec:formulation}
We study the allocation of a fixed post-training budget between SFT and subsequent RL-based training.  
Given a pretrained model $\cM_N$ of size $N$ and a total annotation budget $B$, an allocation ratio $r \in [0, 1]$ assigns $rB$ to SFT and $(1{-}r)B$ to the RL stage. 
The two stages are applied sequentially (SFT followed by RL), yielding a post-trained model that is evaluated using a non-negative, task-specific performance metric $\cP(N, r, B; \tau)$. 
We suppress the task $\tau$ when clear from context.
Following~\citet{raghavendra2025balancing}, we measure $B$ in terms of annotated training samples, assuming comparable annotation costs across SFT and RL; we generalize to heterogeneous annotation costs in \cref{sec: money}. 
We assume training prompts are available for both stages and omit infrastructure and engineering costs.
We simplify the budget and only consider annotated samples, following standard convention~\citep{raghavendra2025balancing, shen2026exploring} and, crucially, annotation dominates the total cost in post-training. 
Using our measured runtimes and current cloud GPU prices in L40s\footnote{\url{https://www.coreweave.com}}, per-example training compute ranges from \$$2\times10^{-5}$ (1B SFT, the cheapest) to \$$3\times10^{-4}$  (8B DPO). Against human annotation (\$$0.5\text{--}1.0$ per example~\citep{kiela2021dynabench, lee2023rlaif}), this is $3\text{--}5$ orders of magnitude smaller. 
Against cheap synthetic annotation (\$$10^{-3}$/example), the margin is smaller, but compute remains below annotation, so annotation is the dominant term and our first-order budget axis.

\paragraph{From point optimum to near-optimal region.}
A natural choice is to optimize the allocation ratio by aiming at the point of optimum, defined as:
\eq{
    \label{eqn:point_optimum}
    r^*(N, B) \;=\; \arg\max_{r \in [0, 1]} \cP(N, r, B).
}
In practice, post-training performance is sensitive to the task, model family, and training method~\citep{caballero2023broken, mckenzieinverse, zhang2024when, schaeffer2025why}, making $r^*$ ill-conditioned and a poor target. 
As we show in \cref{sec: epsilon-region}, $\cP(N, \cdot, B)$ often forms a wide plateau near its maximum, so the question of practical interest is not the single best ratio but the set of ratios achieving near-optimal performance. 
We therefore study the $\epsilon$-near-optimal region, defined as:
\al{
    \label{eqn:optimal_region}
    \cR_\epsilon(N, B) =& \Big\{\, r \in [0, 1]:
    \cP(N, r, B) \geq \nonumber \\
    &\quad(1-\epsilon)\!\max_{r' \in [0, 1]}\!
    \cP(N, r', B) \,\Big\} \ ,
}
where $\epsilon \in [0, 1]$ is a relative performance tolerance (e.g., $\epsilon = 5\%$ allows any ratio retaining at least $95\%$ of peak performance). 
\cref{eqn:optimal_region} reduces to the point-optimum formulation in \cref{eqn:point_optimum} as $\epsilon \to 0$.

\paragraph{Positioning.}
Prior work observes broad trends in allocation ratio $r^*$ across SFT and preference optimization~\citep{raghavendra2025balancing}. However, it does not characterize \emph{where} near-optimal allocations lie, \emph{how wide} the near-optimal region is, or \emph{whether} allocations transfer across model scales. 
To fill this gap, we study how $\cR_\epsilon$ varies with model size, establishing that (i)~the region widens at larger scales under a fixed tolerance, reflecting greater flexibility in practice; and (ii)~near-optimal regions transfer more reliably across scales than exact optima, offering a robust proxy-to-target allocation mechanism.

\subsection{Supervised Fine-Tuning (SFT)}
\label{sec:sft}
SFT adapts a pretrained language model on curated (prompt, response) pairs. 
Given a dataset $\cD_{\mathrm{SFT}} = \{(x_i,y_i)\}_{i=1}^{N_\mathrm{S}}$ of $N_\mathrm{S}$ demonstration samples, where $y_i$ denotes the response for prompt $x_i$, the model is trained via a next-token prediction loss:
\eqs{
    \cL_{\mathrm{SFT}}(\pi_\theta) = -\mathbb{E}_{(x,y)\sim \cD_{\mathrm{SFT}}} \Lb \log \pi_\theta(y \mid x) \Rb,
}
where $\pi_\theta$ denotes the policy parameterized by $\theta$, and $\pi_\theta(y \mid x)$ is the likelihood of the target response $y$ for a given prompt $x$.

\subsection{RL-based Post-Training}
\label{sec:rl}
Beyond supervised imitation, RL fine-tuning equips the model with reward-maximizing or preference-aware behavior. 
We study both the off-policy and on-policy paradigms via Direct Preference Optimization (DPO) and Group Relative Policy Optimization (GRPO), respectively.

\paragraph{Off-policy RL: DPO.}
DPO~\citep{rafailov2023direct} provides a reward-model-free alternative to PPO-based RLHF~\citep{schulman2017proximal}, trained on a static preference dataset $\cD_{\mathrm{DPO}} = \{(x_i, y_{i,w}, y_{i,l})\}_{i=1}^{N_\mathrm{D}}$ of $N_\mathrm{D}$ samples, where $y_{i,w}$ and
$y_{i,l}$ denote the preferred and rejected responses for prompt $x_i$. 
DPO optimizes the policy relative to a fixed reference model $\pi_{\mathrm{ref}}$ using the objective:
\eqs{
    \cL_{\mathrm{DPO}}(\pi_\theta; \pi_{\mathrm{ref}}) \hspace{-0.05mm}=\hspace{-0.05mm} -\mathbb{E}_{(x,y_w,y_l)\sim \cD_{\mathrm{DPO}}} \Lb \log \sigma\left(\beta\Delta\right)\Rb,
}
where
$
    \Delta = \log\frac{\pi_\theta(y_w \mid x)} {\pi_{\mathrm{ref}}(y_w \mid x)} - \log \frac{\pi_\theta(y_l \mid x)} {\pi_{\mathrm{ref}}(y_l \mid x)},
$
$\beta$ controls the strength of the KL penalty, and $\sigma(\cdot)$
is the sigmoid function. 
Since DPO learns from a static preference dataset without sampling new responses during training, it is an off-policy method.

\paragraph{On-policy RL: GRPO.}
GRPO~\citep{shao2024deepseekmath} is an on-policy RL method that updates the model using responses sampled from the current policy. For each prompt $x$, the model generates a group of $N_\mathrm{G}$ responses $\{y_i\}_{i=1}^{N_\mathrm{G}}$, each assigned reward $r_i$. 
GRPO normalizes rewards within the group to compute relative advantages:
$
    A_i = \nicefrac{\Lp r_i - \mathrm{mean}(\{r_j\}_{j=1}^{N_\mathrm{G}}) \Rp}{\mathrm{std}(\{r_j\}_{j=1}^{N_\mathrm{G}})}.
$
The policy is then optimized via a PPO-style clipped objective:
\eqs{
    \cL_{\mathrm{GRPO}}(\pi_\theta) = -\mathbb{E} \Lb \min \Lp \rho_i A_i, \xi_i \Rp \Rb, 
}
where $\xi_i = \mathrm{clip}(\rho_i, 1-\varepsilon_{\mathrm{clip}}, 1+\varepsilon_{\mathrm{clip}}) A_i$,
$
    \rho_i = \nicefrac{\pi_\theta(y_i \mid x)}{\pi_{\theta_{\mathrm{old}}}(y_i \mid x)}
$
is the importance sampling ratio, and $\varepsilon_{\mathrm{clip}}$ is the clipping threshold. 
Unlike DPO, GRPO continuously samples responses from the current policy, making it an on-policy method.

    \section{Analysis}
    \label{sec:analysis}

\subsection{Experimental Setup}
\paragraph{Models and tasks.}
We study post-training allocation using the Llama~3~\citep{grattafiori2024llama},  Qwen~2.5~\citep{yang2024qwen2}, and Qwen~3~\citep{yang2025qwen3} model families across multiple scales up to 14B parameters, evaluating four representative post-training capabilities: math~\citep{cobbe2021training}, instruction following~\citep{zhou2023instruction}, summarization~\citep{stiennon2020learning}, and helpfulness~\citep{wang2024helpsteer1, wang2024helpsteer}.
These tasks cover a diverse range of language model behaviors: reasoning, controllable generation, alignment to user intent, and preference-sensitive response quality. 
An overview is given in \cref{tab:dataset_setting}; further details are in \cref{app:data}. 
Scaling results up to 14B parameters are deferred to
\cref{app:extended-scale}.
\begin{table*}[!ht]
    \centering
    \scriptsize
    \setlength{\tabcolsep}{3pt}
    \renewcommand{\arraystretch}{1.05}
    \begin{tabularx}{\textwidth}{p{0.13\textwidth} X p{0.35\textwidth} p{0.23\textwidth}}
    \toprule
    \textbf{Task} & \textbf{SFT-stage} & \textbf{RL-stage} & \textbf{Evaluation} \\
    \midrule
    Math
    & GSM8K \citep{cobbe2021training}
    & T\"{u}lu3 Grade School Math \citep{lamberttulu3}
    & GSM8K Test Accuracy \\
    
    Instruction following
    & T\"{u}lu3 Persona IF \citep{lamberttulu3}
    & T\"{u}lu3 Persona IF / T\"{u}lu3 RLVR IF \citep{lamberttulu3}
    & IFEval Accuracy \citep{zhou2023instruction} \\
    
    Summarization
    & Reddit TL;DR \citep{volske-etal-2017-tl}
    & Reddit Comparison \citep{stiennon2020learning}
    & ROUGE-L F1\\
    
    Helpfulness
    & HelpSteer \citep{wang2024helpsteer1}
    & HelpSteer2 \citep{wang2024helpsteer}
    & Reward Model \citep{yang2024rewards} \\
    
    \bottomrule
    \end{tabularx}
    \caption{
        Overview of task-specific datasets and evaluation settings. For most tasks, DPO and GRPO use the same or closely matched prompt distributions. For instruction following, GRPO uses the T\"{u}lu3 RLVR data. 
    }
    \label{tab:dataset_setting}
    \vspace{-2mm}
\end{table*}

\paragraph{Allocation and budget grids.}
\label{sec:post-training-allocation}
We instantiate the formulation of \cref{sec:formulation} on discrete grids of allocation ratios and budget values.
Following~\citet{raghavendra2025balancing}, we evaluate over the allocation grid $\cG = \{0.00, 0.25, 0.50,0.75, 1.00\}$\footnote{
To verify that $\cG$ is sufficient, we repeat the 
\cref{sec: scaling-law} analysis on Math and Summarization using a denser $9$-point grid formed by inserting midpoints $\{0.125, 0.375, 0.625,$ $0.875\}$ into $\cG$; results are qualitatively consistent (\cref{app:dense-grid}).
} over budgets $\cB \subset [0, 15\text{k}]$ across different model sizes $\cN$.
For each $(N, r, B) \in \cN \times \cG\times \cB$, we train on $\lceil rB \rceil$ SFT samples followed by $\lfloor (1{-}r)B \rfloor$ RL-stage samples and evaluate the resulting model on $\cP$. 
Performance curves in the small-budget regime ($B < 5\text{k}$) are noisy and occasionally non-monotonic, reflecting instability in early post-training dynamics before convergence. 
We therefore restrict our analysis of scaling behavior to $B \geq 5\text{k}$, consistent with the common practice of excluding pre-convergence points from scaling-law analysis~\citep{kaplan2020scaling, hoffmann2022training}. 
Full performance curves across all budgets, including the pre-convergence regime, are reported in \cref{app:curves}.

\paragraph{Estimating near-optimal region.}
Since $\cP$ is measured on $\cG$, we use the grid-restricted estimate:
\al{
    \label{eqn:grid_restricted}
    \widehat{\cR}_\epsilon(N, B) =& \Big\{\, r \in \cG: 
    \cP(N, r, B) \,\geq\, \nonumber \\
    &\quad (1-\epsilon)\!\max_{r' \in \cG}\!
    \cP(N, r', B) \,\Big\}.
}
We report two complementary grid-based estimators of the width 
of $\widehat{\cR}_\epsilon(N, B)$:

\begin{itemize}[leftmargin=1em]
    \setlength\itemsep{0.01em}
    \item \textbf{Range width:}
    $w_\epsilon^{\,\mathrm{rng}}(N, B) =
    \max \widehat{\cR}_\epsilon - \min \widehat{\cR}_\epsilon$, the span of near-optimal ratios on $[0, 1]$; 
    
    \item \textbf{Count width:}
    $w_\epsilon^{\,\mathrm{cnt}}(N, B) =
    |\widehat{\cR}_\epsilon|/|\cG|$, the fraction of grid points in $\cG$ 
    that are near-optimal.
\end{itemize}
Both estimators lie in $[0, 1]$ and share the same interpretation: the fraction of allocation ratios achieving flexibility under tolerance $\epsilon$, but measure it differently. 
They differ only when $\widehat{\mathcal{R}}_\epsilon$ is non-contiguous on the grid: range width captures the convex-hull span (sensitive to boundary ratios), while count width captures cardinality (sensitive to gaps). 
Empirically, the two yield consistent qualitative trends (\cref{sec: scaling-law}, \cref{app:count}); per-ratio hit-rate heatmaps (\cref{sec: scaling-law}) confirm that near-optimal ratios are generally contiguous, indicating $\widehat{\cR}_\epsilon$ is approximately an interval. 
We report range width $w_\epsilon^{\,\mathrm{rng}}$ in \cref{sec: epsilon-region,sec: scaling-law,sec: money} and count-based analysis in \cref{app:count}.
For model-scale analyses in \cref{sec: epsilon-region,sec: scaling-law,sec: money}, statistics are averaged over budgets in the stable regime ($B \geq 5\text{k}$).

\paragraph{Training Setup.}
For RL-based post-training, we primarily analyze off-policy preference optimization (DPO) for training efficiency, and extend to on-policy optimization (GRPO) with verifiable rewards. 
To ensure fair comparisons across allocation ratios, SFT and RL-stage training data for each task come from a consistent source: either human-annotated or machine-generated.
Given the large number of post-training runs required, we use LoRA~\citep{hu2022lora} for training efficiency. 
Detailed hyperparameters are in \cref{app:hyper}.

\subsection{Sensitivity of Budget Allocation}
\label{sec: epsilon-region}
As established in \cref{sec:formulation}, our analysis centers on the $\epsilon$-near-optimal region $\cR_\epsilon(N, B)$ rather than the point optimum $r^*(N, B)$. 
Throughout this section and the appendix, $\cR_\epsilon$ refers to the grid-restricted width estimate of $\widehat{\cR}_\epsilon(N, B)$ (\cref{eqn:grid_restricted}), and we report the range-based width ${w}_\epsilon^{\,\mathrm{rng}}$ as the primary width statistic. 
We first characterize how wide $\cR_\epsilon$ is across tasks and model families.

\begin{figure}[!ht]
    \centering
    \includegraphics[width=\columnwidth]{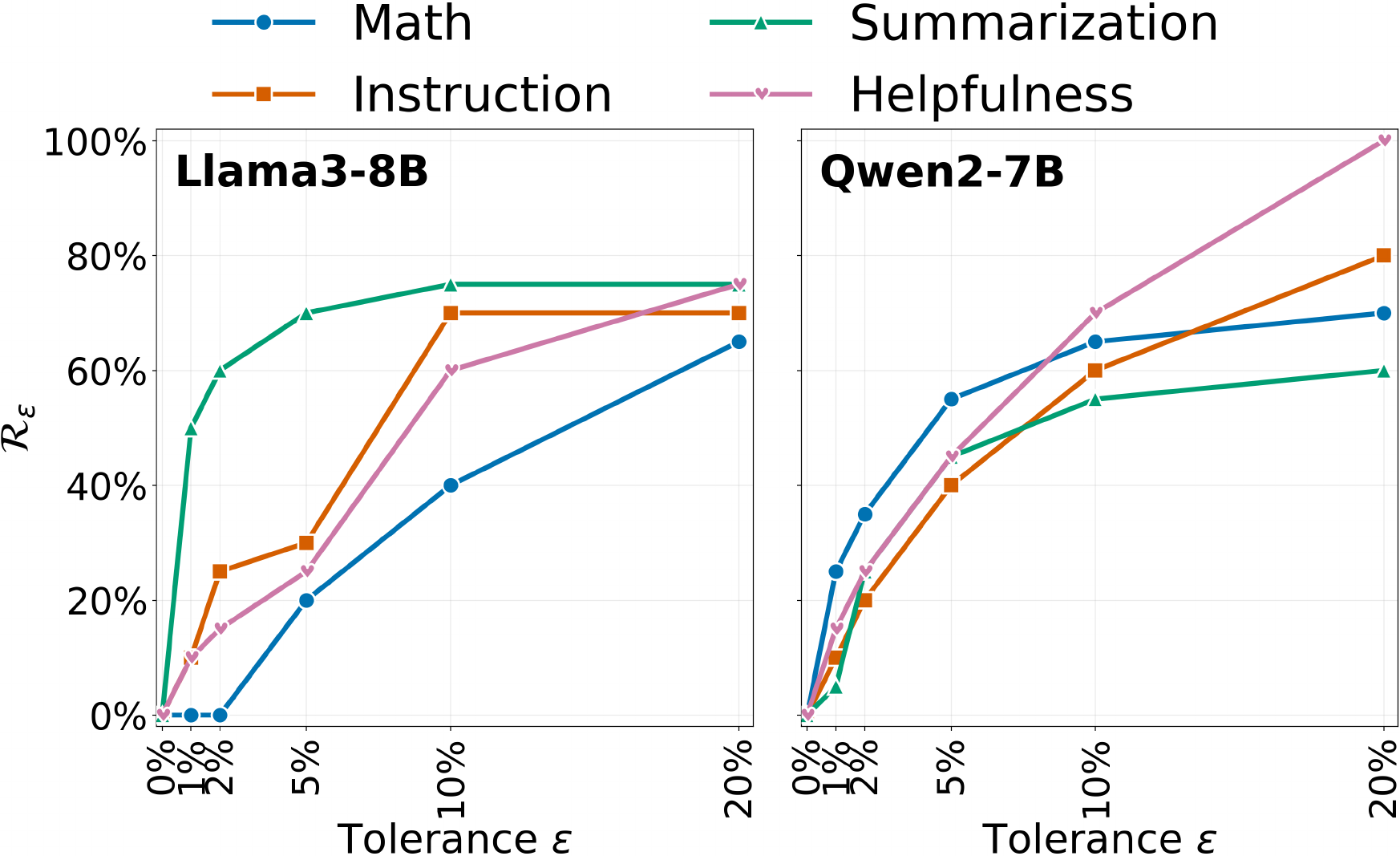}
    \caption{
        Tolerance $\epsilon$ versus near-optimal-region width $\cR_\epsilon (={w}_\epsilon^{\,\mathrm{rng}})$ for Llama3-8B and Qwen2.5-7B across four tasks. The near-optimal region expands consistently with tolerance across tasks and model families.
    }
    \label{fig:tolerance-width}
    \vspace{-3mm}
\end{figure}

\paragraph{Optimal allocation as a near-optimal region rather than a single ratio.} 
As shown in \cref{fig:tolerance-width}, the near-optimal region width $\cR_\epsilon$ expands consistently with tolerance across tasks and model families.
Under a $10\%$ tolerance, retaining at least $90\%$ of optimal performance, most tasks admit near-optimal ratios covering $55\%$--$75\%$ of the feasible allocation space. 
Post-training performance is therefore broadly insensitive to allocation choice within a wide neighborhood, rather than being concentrated at a single ratio. 
This empirically supports the plateau structure of $\cP(N, \cdot, B)$ posited in \cref{sec:formulation} and motivates studying $\cR_\epsilon$.  

\subsection{Scaling Behavior of Budget Allocation}
\label{sec: scaling-law}
\begin{figure*}[!ht]
    \centering
    \includegraphics[width=\textwidth]{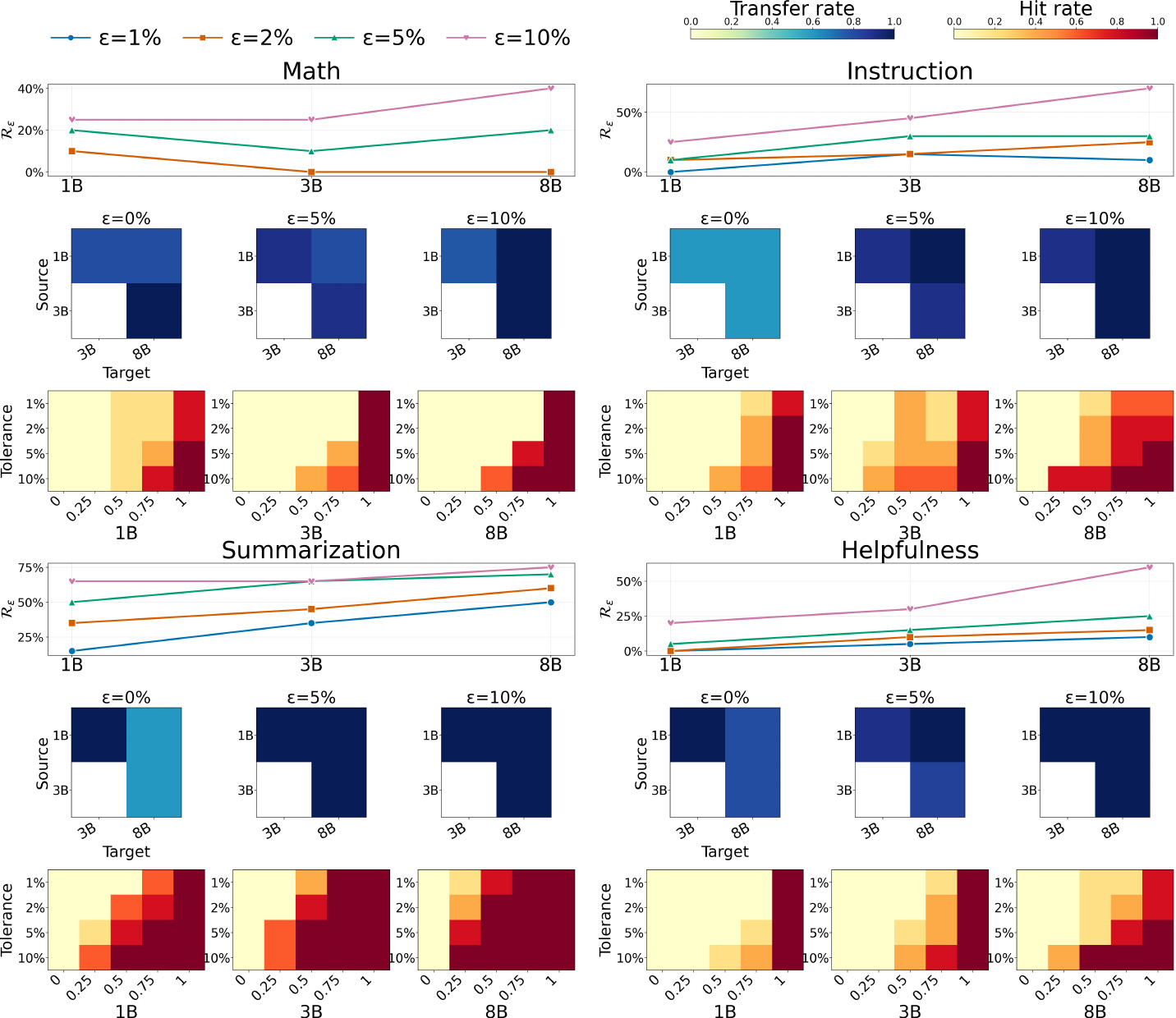}
    \caption{
        \textbf{Main analysis for the Llama family with SFT--DPO training.} Each task panel has three rows. 
        \textbf{Row 1 (line plot):} mean near-optimal region grid-restricted estimates (for brevity, we use near-optimal region) width versus model size $N$ per tolerance level $\epsilon$; width generally increases with model scale. 
        \textbf{Row 2 (transfer matrices):} cross-size overlap of near-optimal region $\widehat{\cR}_\epsilon(N, B)$ (small $\rightarrow$ large); these regions transfer more reliably across model scales than point optimum. 
        \textbf{Row 3 (heatmaps):} per-ratio hit rate across budgets; each (tolerance, ratio) cell indicates the fraction of budgets $B$ for which that ratio lies within the near-optimal region.
    }
    \label{fig:main-llama-width}
    \vspace{-5mm}
\end{figure*}

\begin{figure*}[!ht]
    \centering
    \includegraphics[width=\textwidth]{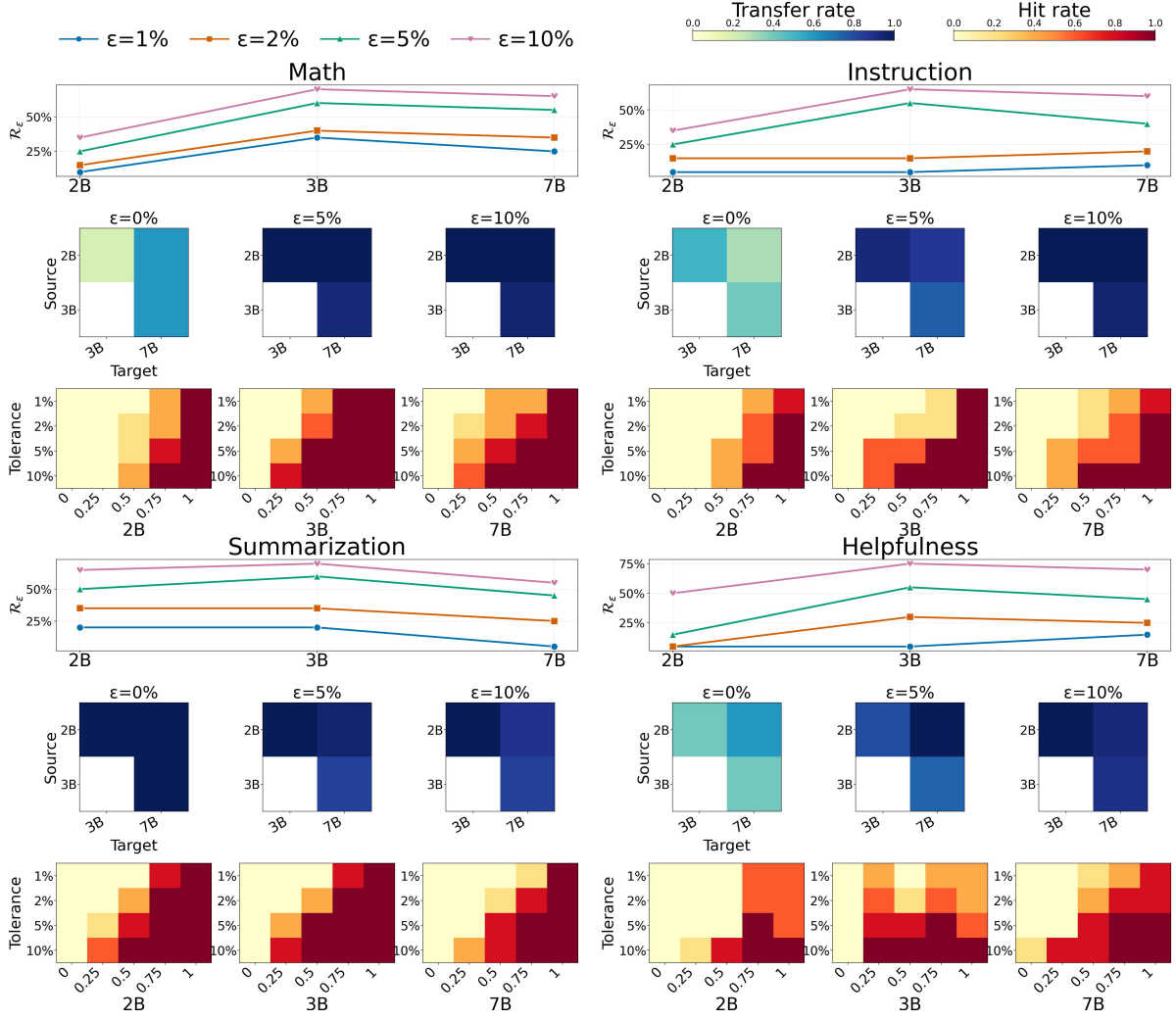}
    \caption{
    \textbf{Main analysis for Qwen~2.5 family with SFT--DPO training stages.}
    Qualitative trends align with \cref{fig:main-llama-width}, supporting the same findings. 
    For brevity, here and throughout all figures involving the Qwen2.5 family, 2B denotes the Qwen2.5-1.5B model.
    }
    \label{fig:main-qwen-width}
    \vspace{-3mm}
\end{figure*}

\begin{figure*}[!ht]
    \centering
    \includegraphics[width=\textwidth]{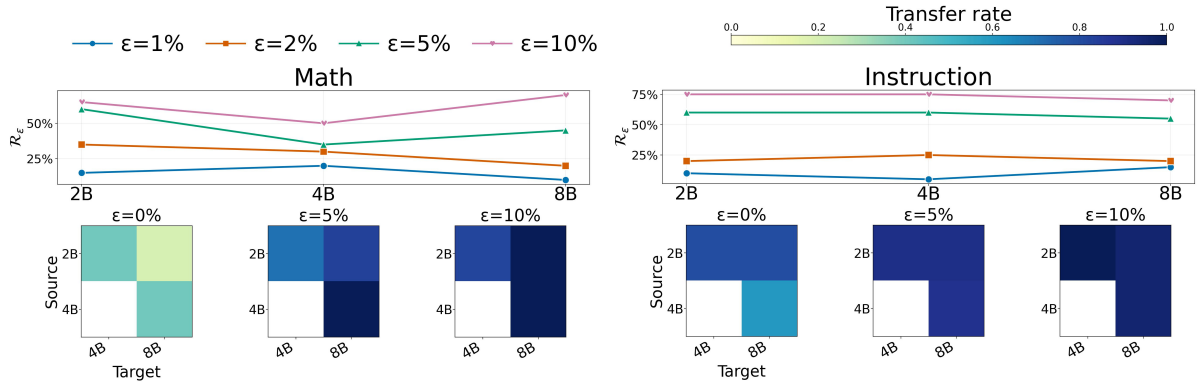}
    \caption{
    \textbf{Main analysis for Qwen~3 family with SFT--DPO training stages.}
    Qualitative trends align with \cref{fig:main-llama-width}, supporting the same findings.
    For a more complete figure with a heat map, see \cref{app:complete}. 
    }
    \label{fig:main-qwen3-width}
    \vspace{-5mm}
\end{figure*}

Having established that $\cR_\epsilon$ is substantially wider than the point optimum across all tasks, we now study how it scales with model size $N$ and whether near-optimal allocations transfer from a small proxy model $M_{N_s}$ to a larger target model $M_{N_t}$. 
Results are shown in \cref{fig:main-llama-width,fig:main-qwen-width,fig:main-qwen3-width}.

\paragraph{Near-optimal region widens with model scale.}
The first row of each sub-figure reports the average width  ${w}_\epsilon^{\,\mathrm{rng}}(N, B)$ across budgets.  
Across tasks and model families, the near-optimal region under a fixed tolerance $\epsilon$ generally widens with $N$: smaller models often admit only a narrow set of near-optimal ratios under strict tolerances, whereas larger models accommodate a broader range.  
From a practitioner's view, deploying a larger model thus affords greater freedom in choosing $r$ while retaining a target fraction of peak performance.

This widening can be driven by larger models achieving higher absolute peak performance $\cP^*(N, B)$: under a fixed tolerance $\epsilon$, a higher peak implies more absolute slack at the same threshold  $(1 - \epsilon)\cP^*$, which mechanically admits more ratios into  $\cR_\epsilon$. 
A full decomposition (into peak-performance scaling versus intrinsic changes in allocation sensitivity) is task- and method-dependent; a detailed analysis in \cref{app:abs-tolerance} identifies edge cases in which the widening can fail under an alternative definition of tolerance. 
We therefore interpret the widening as a \emph{fixed-tolerance phenomenon} rather than as evidence that allocation sensitivity itself decreases with scale.

\paragraph{Near-optimal regions transfer more reliably across model scales than point optimum.}
To quantify cross-scale transferability, we measure the fraction of near-optimal allocations identified on a small proxy model $M_{N_s}$ that remain near-optimal on a larger target model $M_{N_t}$:
\eqs{
    T_\epsilon(N_s \!\to\! N_t;\, B) = \frac{|\widehat{\cR}_\epsilon(N_s, B) \cap \widehat{\cR}_\epsilon(N_t, B)|}{|\widehat{\cR}_\epsilon(N_s, B)|}.
}
The asymmetry is intentional: we care about preserving near-optimal allocations from a cheaper proxy as we scale up. 
We compute $T_\epsilon$ \emph{per-budget} $B$, and aggregated across budgets for reporting. The second row of each sub-figure reports the budget-aggregated $T_\epsilon$ for each model-size pair.

Under point optimum ($\epsilon = 0$), transferability is sometimes high but can be inconsistent across tasks, budgets, and model families, particularly for the Qwen2.5 family (\cref{fig:main-qwen-width}).
Allowing a modest tolerance substantially and consistently improves transfer, indicating that near-optimal regions are a more robust object for cross-scale transfer than the exact point optimum. 
A wider region on the target model mechanically increases its overlap with the source region, providing a structural explanation for why  $T_\epsilon$ improves as $\epsilon$ increases.

The third row of each sub-figure jointly illustrates the expansion and transferability of near-optimal regions. 
Each heatmap reports, across budgets, how frequently each allocation ratio falls within $\cR_\epsilon$ at a given $(N, \epsilon)$. Darker cells indicate ratios that consistently achieve near-optimal performance. 
As $\epsilon$ increases, darker regions expand \emph{contiguously}: neighboring allocation ratios generally yield comparable performance, supporting the contiguity assumption underlying our width estimator (\cref{sec:post-training-allocation}). 
Moreover, the high-frequency regions of smaller models typically overlap with, or are contained within, those of larger models, further supporting cross-scale transfer.

We note one apparent exception: for Qwen~2.5 summarization, $\cR_\epsilon$  narrows marginally at larger scales under higher tolerances. 
This reflects a potential saturation effect (smaller Qwen models for summarization already admit wide, near-optimal regions, leaving little room for further widening) rather than a collapse in allocation behavior. 
Cross-scale transfer for Qwen~2.5 summarization remains strong (transfer matrices and hit-rate heatmaps, \cref{fig:main-qwen-width}), consistent with this interpretation.  
Together, these findings indicate that smaller proxy models can reliably guide post-training allocation for larger target models without requiring expensive large-scale allocation sweeps. 
The near-optimal region $\cR_\epsilon$ is a more robust objective for cross-scale transfer than the point optimum $r^*$, with practical implications for the cost-efficient post-training pipeline design.

\begin{figure*}[!ht]
    \centering
    \includegraphics[width=\textwidth]{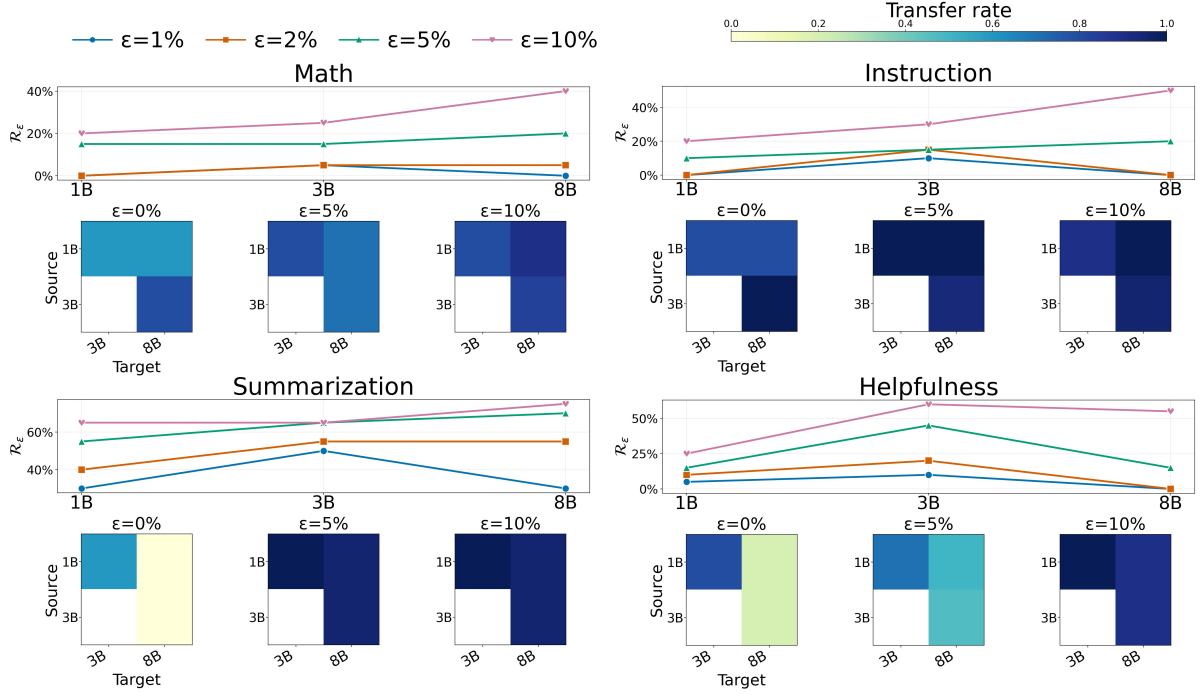}
    \caption{
    \textbf{Analysis for Llama family with SFT--GRPO training stages. }
    Qualitative trends align with \cref{fig:main-llama-width}, supporting the same findings.
    For a more complete figure with a heat map, see \cref{app:complete}. 
    }
    \label{fig:llama-rlvr-width} 
    \vspace{-2mm}
\end{figure*}

\paragraph{Trends extend to on-policy post-training.}
The preceding analyses use DPO, an off-policy preference optimization method chosen for training efficiency. 
We now extend the analysis to on-policy RL with GRPO trained on verifiable rewards or reward models.  
\cref{fig:llama-rlvr-width} reports results for the SFT--GRPO setting on the Llama family across all four tasks. The qualitative pattern matches the SFT--DPO setting: $\cR_\epsilon$ widens with tolerance, and near-optimal regions transfer across model scales more reliably than the point optimum.  
Compared with SFT--DPO, the SFT--GRPO results exhibit slightly stronger fluctuations, particularly on Helpfulness, consistent with the higher variance of on-policy training, where policy sampling, reward estimation, and iterative updates introduce additional noise. Our conclusion holds across both settings: the near-optimal region $\cR_\epsilon$ is a more reliable object for cross-scale transfer than the point optimum $r^*$, for both off-policy and on-policy post-training.

\subsection{Extension to Cost Asymmetry Setting}
\label{sec: money}

\paragraph{Allocation under heterogeneous annotation costs.} 
Previous sections measure post-training budgets by the number of annotated samples, implicitly assuming equal annotation costs for SFT and RL-stage data. 
In practice, high-quality supervised demonstrations are typically more expensive than preference annotations or verifier-based rewards~\citep{raghavendra2025balancing}; for instance, generating synthetic SFT data via frontier LLMs can cost roughly $1$--$2\times$ as much as preference-style annotations under contemporary API pricing~\citep{wang2023self, lee2023rlaif}. 
We examine if our scaling and transfer trends persist when SFT and RL data incur different annotation costs, focusing on settings where DPO is the RL-stage objective.

\paragraph{Experimental setting.}
Fix a monetary budget $B$ (in dollars) and let $\rho := c_{\mathrm{SFT}} /  c_{\mathrm{DPO}}$ denote the SFT-to-DPO cost ratio. 
An allocation $r$ assigns $rB$ dollars to SFT and $(1-r)B$ to DPO,  yielding $n_{\mathrm{SFT}} = rB / c_{\mathrm{SFT}}$ and  $n_{\mathrm{DPO}} = (1-r)B / c_{\mathrm{DPO}}$ samples.  
We consider $\rho \in \{1, 2, 5, 10\}$, where larger $\rho$ corresponds to progressively more expensive SFT annotations.  
Following \citet{raghavendra2025balancing}, we set $c_{\mathrm{DPO}} =  \$0.001$ and scale $c_{\mathrm{SFT}}$ accordingly; $\rho = 1$ recovers the sample-count budget of \cref{sec:post-training-allocation}. 
We focus on $\rho \geq 1$ because SFT annotations (human-written or LLM-distilled responses) are typically more expensive than preference pairs or verifier-based rewards in contemporary post-training pipelines.  
We exclude $r = 0$ as it consistently underperforms other ratios and requires disproportionately more data when $\rho > 1$. 
We primarily analyze the Math task; additional results across tasks are deferred to \cref{app:cost}.

\begin{figure*}[!ht]
    \centering
    \includegraphics[width=\textwidth]{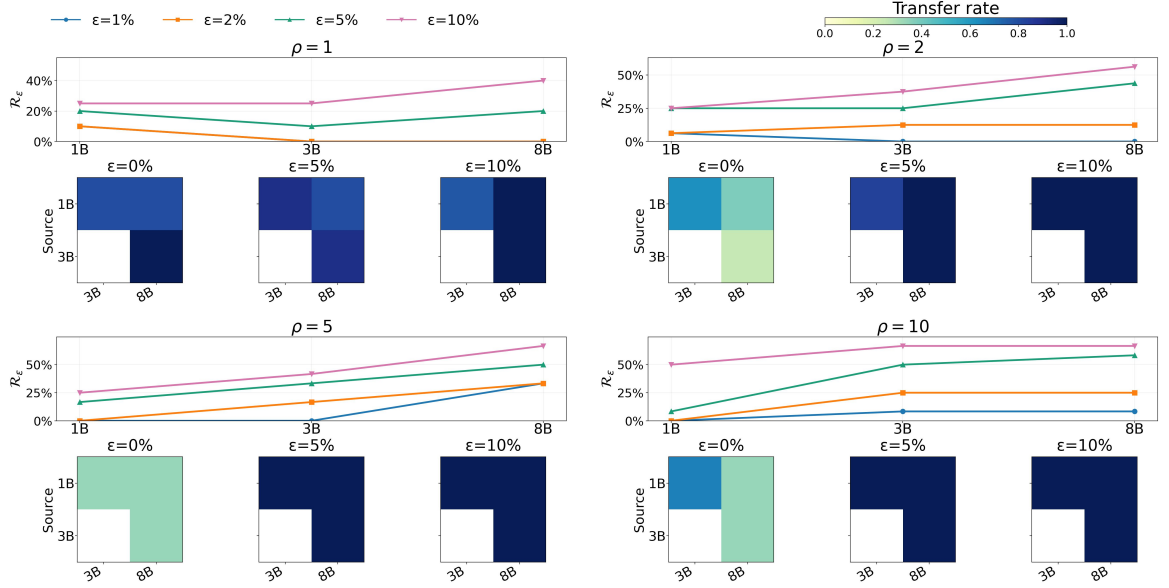}
    \caption{
        \textbf{Effect of heterogeneous annotation costs on transferable allocation regions for the math task.}
        We simulate varying SFT-to-RL annotation cost ratios across the Llama model family. Qualitative trends remain consistent with the equal-cost setting: larger tolerance thresholds produce broader near-optimal regions and stronger cross-scale transferability. A more complete figure with heatmaps is in \cref{app:complete}.
    }
    \label{fig: cost_math} 
    \vspace{-4mm}
\end{figure*}

\paragraph{Scaling trends persist under heterogeneous annotation costs.}
\cref{fig: cost_math} shows that the qualitative scaling behavior remains largely unchanged under heterogeneous annotation costs. 
Across $\rho$ settings, larger tolerance thresholds continue to produce wider  $\cR_\epsilon$ and stronger cross-scale transfer $T_\epsilon$.
Additionally, as $\rho$ increases, $\cR_\epsilon$ generally widens at the same tolerance: a wider range of allocations achieves comparable performance when SFT annotations are relatively expensive.
From a practitioner's perspective, this indicates \emph{increased flexibility} in choosing the SFT--DPO budget split as SFT becomes more costly: the choice of $r$ matters less, and other criteria (annotation logistics, sample reuse) can drive the final allocation. 
Overall, near-optimal allocation regions remain transferable and stable across a range of plausible cost asymmetries between SFT and DPO  annotations, supporting the use of small-proxy allocation guidance in cost-constrained pipelines.

\vspace{-0.65mm}
\subsection{From Analysis to Practice: A Proxy Allocation Procedure}
\label{sec:procedure}
\vspace{-0.5mm}
The preceding results suggest a simple recipe: use\\ 
a small proxy model to identify a robust allocation region, choose a tolerance that reflects the desired trade-off between performance and allocation flexibility, and then select a ratio from that region. We describe this procedure as follows:
\begin{enumerate}
    \item \textbf{Proxy sweep.} Take the smallest model in the target family as the proxy $M_{N_s}$ and evaluate the 5-point ratio grid $\cG = \{0, 0.25, 0.5, 0.75, 1\}$ at the \emph{target} budget $B$. Five points are sufficient: a denser $9$-point grid identifies the same regions (\cref{app:dense-grid}).

    \item \textbf{Choose the tolerance.} Set $\varepsilon$ according to how much peak performance the practitioner is willing to trade for allocation flexibility; i.e., retain at least $1-\varepsilon$ of the proxy's peak performance. We recommend $\varepsilon \in [5\%, 10\%]$, a range in which proxy transfer is reliable across all families, tasks, and budgets we evaluate.
    
    \item \textbf{Select the ratio.} Use the proxy sweep to form the $\varepsilon$-optimal region $\widehat{\cR}_\varepsilon(N_s, B) = [\min \widehat{\cR}_\varepsilon, \max \widehat{\cR}_\varepsilon]$, and choose its midpoint. If the final run must use one of the evaluated ratios, use the nearest grid point instead. The midpoint provides a robust choice for transfer: near-optimal ratios form contiguous regions (\cref{sec: scaling-law}, \cref{app:count}), so the midpoint is less sensitive to small shifts in the optimal region between the proxy and target models than a boundary ratio or the exact proxy optimum.
\end{enumerate}

We test this three-step procedure using our existing 1B--8B runs across all tasks and both the Llama and Qwen~2.5 families. The procedure identifies a target-region ratio that satisfies the chosen tolerance in $94.3\%$ of cases for $\varepsilon=5\%$ and $97.1\%$ of cases for $\varepsilon=10\%$. These results support the proxy sweep as a practical approach to selecting robust allocation ratios without exhaustive tuning at the target scale.
As post-training scaling can be fragile, exhibiting phenomena such as broken or inverse scaling, metric discontinuities, and cross-scale rank inversions \cite{caballero2023broken, mckenzieinverse, schaeffer2025why}, we focus on understanding the mechanisms underlying this transfer and identifying when the above recipe is reliable, rather than proposing a universal scaling law.

\vspace{2mm}

    \section{Related Work}
    \label{sec:related_work}

\paragraph{Budget allocation between training stages.}
The closest line of work to ours is \citet{raghavendra2025balancing}, who study the trade-off between SFT and preference optimization under fixed annotation budgets and observe a generic trend in the optimal SFT--DPO ratio as a function of total budget: SFT dominates in the low-data regime while preference optimization becomes more favorable at scale. 
Their analysis identifies this broad trend but does not characterize the optimal-allocation region or its transferability across model scales. 
\citet{li2026getting} study an orthogonal question: given a fixed SFT data budget, when to stop SFT for best downstream RL performance. 
They propose Adaptive Early-Stop Loss (AESL) and show that diversity-based early stopping (e.g., entropy, self-BLEU) outperforms performance-based stopping criteria due to distributional drift during SFT. 
Their decision variable is the SFT \emph{stopping step} at a fixed $n_{\text{SFT}}$; ours is the SFT \emph{budget share} relative to RL. The two questions are complementary.

A broader set of works examines budget trade-offs between other training stages: pretraining vs. fine-tuning~\citep{bai2021pre}, finetuning vs. distillation from a larger teacher~\citep{kang2023distill, busbridge2025distillation}, and the interplay between SFT and RL methods, including DPO--PPO comparisons~\citep{ivison2024unpacking}, forgetting~\citep{fernando2024understanding}, generalization~\citep{kirk2024understanding}, and alignment trade-offs~\citep{saeidi2025insights}. 
A complementary line of work treats SFT and DPO as connected via an implicit reward formulation~\citep{wang2026implicit}, which motivates the budget-allocation question we study. 
Our contribution differs from all of the above works in characterizing not a single optimal ratio but the entire $\epsilon$-near-optimal region and its cross-scale transferability. 
Additional related work on data mixture optimization and cross-scale extrapolation is deferred to \cref{app:related}. 

\vspace{3mm}

    \section{Conclusion}
    \label{sec:conclusion}

We studied the problem of dividing a fixed annotation budget between SFT and RL in LLM post-training, framing it in terms of near-optimality rather than exact optimality. 
Our experiments show that the near-optimal region (i.e., the set of allocations within a fixed performance tolerance of the best observed allocation) is wide even for small tolerances ($2$--$10\%$), widens consistently with model scale, and transfers reliably from small proxy models to large target models, yielding a practical strategy that eliminates the need for exhaustive large-scale grid search. 
These findings hold across tasks, model families, off-policy preference-based methods, and on-policy RL, and extend to settings with cost asymmetry between SFT and RL annotation. 
A limitation of this work is that our analysis is restricted to two-stage SFT$\to$RL pipelines; extending the near-optimality framework to more complex multi-stage or interleaved schedules and to settings where the annotation cost model is uncertain or adaptive is a promising future direction.

    \clearpage
    \section*{Limitations}
    \label{sec:limitation}

\para{Target-task training setting.}
All experiments train on task-specific data drawn from the same distribution as the evaluation, following the convention of~\citet{raghavendra2025balancing}. This in-distribution setting yields comparatively predictable scaling behavior. The out-of-distribution regime, where general-purpose training data (e.g., a broad instruction or preference mix) must transfer to a specific held-out target task, scales less reliably, with larger generalization gaps and frequent
non-monotonicity in performance versus budget~\citep{caballero2023broken, kirk2024understanding}. Whether near-optimal allocation regions and their cross-scale transferability extend to OOD evaluation remains an important open question.
  
\para{Model and data scale ceiling.}
Our analysis covers model scales up to 14B parameters (leaving Llama 3 70B and Qwen 2.5 32B--72B for future verification) and budgets up to $15\text{k}$ examples per task in the majority setting (cost asymmetry may use more training examples). The data ceiling is at or near the publicly available task-specific training corpus for several of our SFT and RL settings: GSM8K provides $\sim$7.5K SFT examples and $\sim$16.6K synthetic preference pairs~\citep{cobbe2021training, raghavendra2025balancing}; HelpSteer~2 contains $\sim$20K examples used for both  stages~\citep{wang2024helpsteer}; and GRPO-IFeval is itself a $\sim$15K-example training set~\citep{lamberttulu3}. Within our tested range, the cross-scale transferability trends are generally consistent. Extending verification to frontier scales would entail substantial data collection and computational costs, which may be beyond the scope of academic budgets.
  
\para{Restricted RL algorithms.}
Other on-policy algorithms, such as PPO, and other off-policy algorithms, such as SimPO~\citep{meng2024simpo}, are not tested. We mainly compare the most popular and generally used RL policies, namely DPO and GRPO, given computational constraints. However, the consistent results across tasks and settings demonstrate the potential of generalization. 

\para{Simplicity of cost model.}
The notion of budget in our setting is relatively simple; we do not consider training compute, GPU time, or the cost of experimental search. Jointly modeling annotation, compute, and search costs into a single scalar budget is non-trivial and lacks precedent in cross-scale scaling laws. The closest studies treat annotation-vs-compute as a two-way dollar trade-off~\citep{bai2021pre, kang2023distill}, without the cross-scale transfer that we study.

    \section*{Ethical Considerations}

This work is an empirical study of the allocation of budget between supervised fine-tuning and RL-based post-training. All experiments use publicly released model families (Llama~3, Qwen~2.5, and Qwen~3) and publicly available task-specific datasets under their original licenses. We conduct no human-subjects research and collect or release no new data or models. The pretrained models inherit any biases or representational issues present in their original training data, which we do not separately audit.
Our experimental sweep involves on the order of hundreds to thousands of post-training runs across the Cartesian product of model sizes, allocation ratios, budgets, tasks, and algorithms. We use parameter-efficient fine-tuning (LoRA) and cap model sizes at 8B parameters per family (except Qwen2.5 14B model in \cref{fig:app-qwen14-width-full}) to make the sweep tractable on academic compute. 
Our practical contribution: that near-optimal SFT--RL allocation regions transfer from small proxy models to larger models, is intended to reduce the compute required for principled post-training decisions at scale, broadening access for academics and practitioners. We do not foresee specific dual-use risks beyond those already associated with publicly released LLMs and standard post-training methods.

    \section*{Acknowledgments}
    \label{sec:acknowledgment}
    This research is supported by the National Research Foundation, Singapore under its National Large Language Models Funding Initiative (AISG Award No: AISG-NMLP-$2024$-$001$).
This research is supported by the National Research Foundation (NRF), Prime Minister’s Office, Singapore under its Campus for Research Excellence and Technological Enterprise (CREATE) programme. The Mens, Manus, and Machina (M3S) is an interdisciplinary research group (IRG) of the Singapore MIT Alliance for Research and Technology (SMART) centre.
Jingtan Wang is supported by the Institute for Infocomm Research of the Agency for Science, Technology and Research (A*STAR).

    \bibliography{references}

    \clearpage
    \newpage
    \appendix
    \crefalias{section}{appendix}
    \crefalias{subsection}{appendix}
    

\startcontents[appendix]
\label{sec:appendix}
\addcontentsline{toc}{section}{Appendix}
\section*{Table of Contents}
\printcontents[appendix]{}{1}{}
\newpage

\section{Experimental Details}
\subsection{Tasks}
\label{app:data}
\paragraph{Dataset construction details.}
Our dataset construction largely follows \citet{raghavendra2025balancing}. 
For each task, we construct an SFT dataset consisting of prompt-response pairs and an RL-stage dataset consisting of either preference pairs for DPO or prompts with reward signals for GRPO. 
We use the same dataset sources, pre-processing rules, and evaluation protocols as in the setup of \citet{raghavendra2025balancing}, unless otherwise stated. 
For DPO, preference data are formatted as triples $(x,y_w,y_l)$, where $y_w$ and $y_l$ denote the preferred and rejected responses. 
For GRPO, we reuse the same prompts $x$ from the DPO data and task-specific evaluation metrics as rewards. Note that for summarization, we use the preferred column's response as the reference answer when calculating the ROUGE-L. 
The only deviation is the instruction-following task, where we use the original T\"ulu3 RLVR instruction-following dataset for GRPO, since constructing evaluation instructions that serve as the reward directly from the preference-dataset prompts is less straightforward. For the evaluation datasets, we follow the standard benchmark splits for GSM8K~\cite{cobbe2021training} and IFEval~\cite{zhou2023instruction}, and use the official test split of the summarization dataset~\cite{volske-etal-2017-tl}. For HelpSteer, we construct a local 200-example held-out test split for evaluation.
Dataset sources and download links are listed in \cref{tab:task_detail}.

\begin{table*}[!ht]
\centering
\small
\begin{tabularx}{\textwidth}{@{}llX@{}}
\toprule
\textbf{Task} & \textbf{Setting} & \textbf{Dataset / Download Link} \\
\midrule

Grade School Math 
& SFT 
& GSM8K generations: \url{https://huggingface.co/datasets/mohit-raghavendra/gsm8k-gpt4ogenerations} \\

& RL
& GSM8K preferences: \url{https://huggingface.co/datasets/mohit-raghavendra/gsm8k-gpt4opreferences} \\

\midrule

Instruction Following 
& SFT 
& T\"ulu3 Instruction Following: \url{https://huggingface.co/datasets/allenai/tulu-3-sft-personas-instruction-following} \\

& DPO 
& T\"ulu3 Instruction Following preferences: \url{https://huggingface.co/datasets/allenai/tulu-3-pref-personas-instruction-following} \\

& GRPO 
& GRPO-IFeval: \url{https://huggingface.co/datasets/allenai/RLVR-IFeval} \\

& Evaluation 
& IFEval: \url{https://github.com/google-research/google-research/tree/master/instruction_following_eval} \\
\midrule
Summarization 
& SFT 
& Reddit TL;DR: \url{https://huggingface.co/datasets/UCL-DARK/openai-tldr-filtered} \\

& RL
& Reddit comparison dataset: \url{https://huggingface.co/datasets/UCL-DARK/openai-tldr-summarisation-preferences} \\

\midrule

Helpfulness 
& SFT 
& HelpSteer1: \url{https://huggingface.co/datasets/nvidia/HelpSteer};
  HelpSteer2: \url{https://huggingface.co/datasets/nvidia/HelpSteer2} \\
& RL
& HelpSteer1: \url{https://huggingface.co/datasets/nvidia/HelpSteer};
  HelpSteer2: \url{https://huggingface.co/datasets/nvidia/HelpSteer2} \\
& Evaluation 
& Reward model: \url{https://huggingface.co/Ray2333/gpt2-large-helpful-reward_model}\\
\bottomrule
\end{tabularx}
\caption{Detailed dataset and evaluation references for reproducibility. 
For each task, we report the training and evaluation setting together with the corresponding public dataset or benchmark source. }
\label{tab:task_detail}
\end{table*}

\begin{table*}[!ht]
\centering
\begin{tabular}{ll}
\hline
\textbf{Parameter} & \textbf{Value} \\
\hline
Learning Rate & $1 \times 10^{-5}$ \\
Effective Batch Size & 16 \\
Number of Epochs & 1 (2 for Instruction tuning) \\
Warmup Ratio & 0.1 \\
Learning Rate Scheduler & cosine \\
Optimizer & adamw\_8bit \\
Weight Decay & 0.01 \\
\hline
\end{tabular}
\caption{SFT Hyperparameter Settings}
\label{tab:sft_hyperparameters}
\end{table*}

\begin{table*}[!ht]
\centering
\begin{tabular}{ll}
\hline
\textbf{Parameter} & \textbf{Value} \\
\hline
Learning Rate & $1 \times 10^{-6}$ \\
Effective Batch Size & 16 \\
Number of Epochs & 1 (2 for Instruction tuning)\\
Warmup Ratio & 0.1 \\
Learning Rate Scheduler & cosine \\
Optimizer & adamw\_8bit \\
Weight Decay & 0.01 \\
$\beta$ & 0.1 \\
\hline
\end{tabular}
\caption{DPO Hyperparameter Settings}
\label{tab:dpo_hyperparameters}
\end{table*}

\begin{table*}[!ht]
\centering
\begin{tabular}{ll}
\hline
\textbf{Parameter} & \textbf{Value} \\
\hline
Learning Rate & $1 \times 10^{-6}$ \\
Effective Batch Size & 16 \\
Number of Epochs & 1 (2 for Instruction tuning)\\
Warmup Ratio & 0.1 \\
Learning Rate Scheduler & cosine \\
Optimizer & adamw\_8bit \\
Weight Decay & 0.01 \\
Number of Generations & 4 \\
Temperature & 0.9 \\
Top-$p$ & 1.0 \\
Top-$k$ & -1 \\
Minimum-$p$ & 0.0 \\
Repetition Penalty & 1.0 \\
\hline
\end{tabular}
\caption{GRPO Hyperparameter Settings}
\label{tab:rlvr_hyperparameters}
\end{table*}

\subsection{Hyper-parameter settings}
\label{app:hyper}
Hyperparameter settings for each stage we used are in \cref{tab:sft_hyperparameters,tab:dpo_hyperparameters,tab:rlvr_hyperparameters}. For all experiments, we use LoRA with rank $r=32$, scaling factor $\alpha=32$, dropout of 0, no bias terms, and target the attention modules. The main experiments are run with seed 42 on L40s and H200s. 

\section{Complete Performance Curves with Different Budgets}
\label{app:curves}
\begin{figure*}[!ht]
    \centering
    \includegraphics[width=\textwidth]{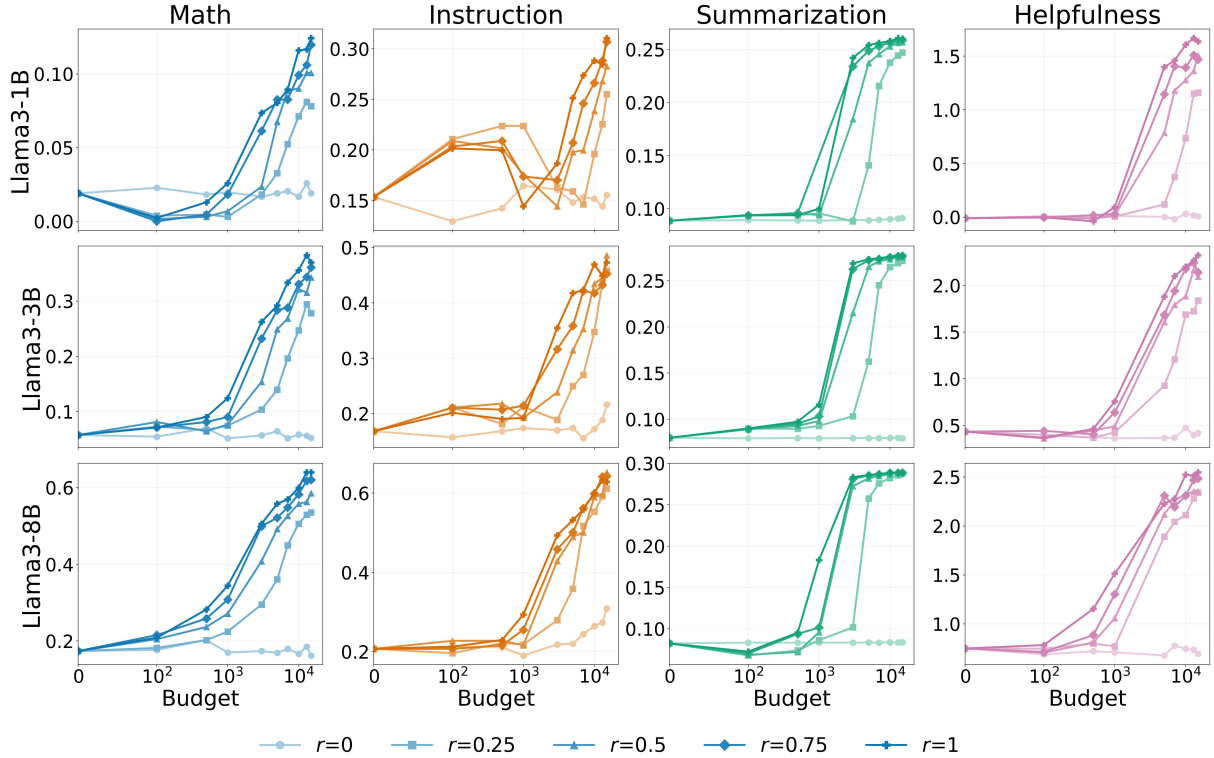}
    \caption{
    Performance versus training budget across SFT--DPO allocation ratios on the Llama 3 model family, organized by model size (rows) and tasks (columns). Curves can be noisy in the small-budget regime ($B < 5\text{k}$) and stabilize for $B \geq 5\text{k}$, where ratios converge to comparable downstream performance: consistent with the broad near-optimal regions in \cref{sec: epsilon-region}. }
    \label{fig:appendix-llama-3x4}
\end{figure*}

\begin{figure*}[!ht]
    \centering
    \includegraphics[width=\textwidth]{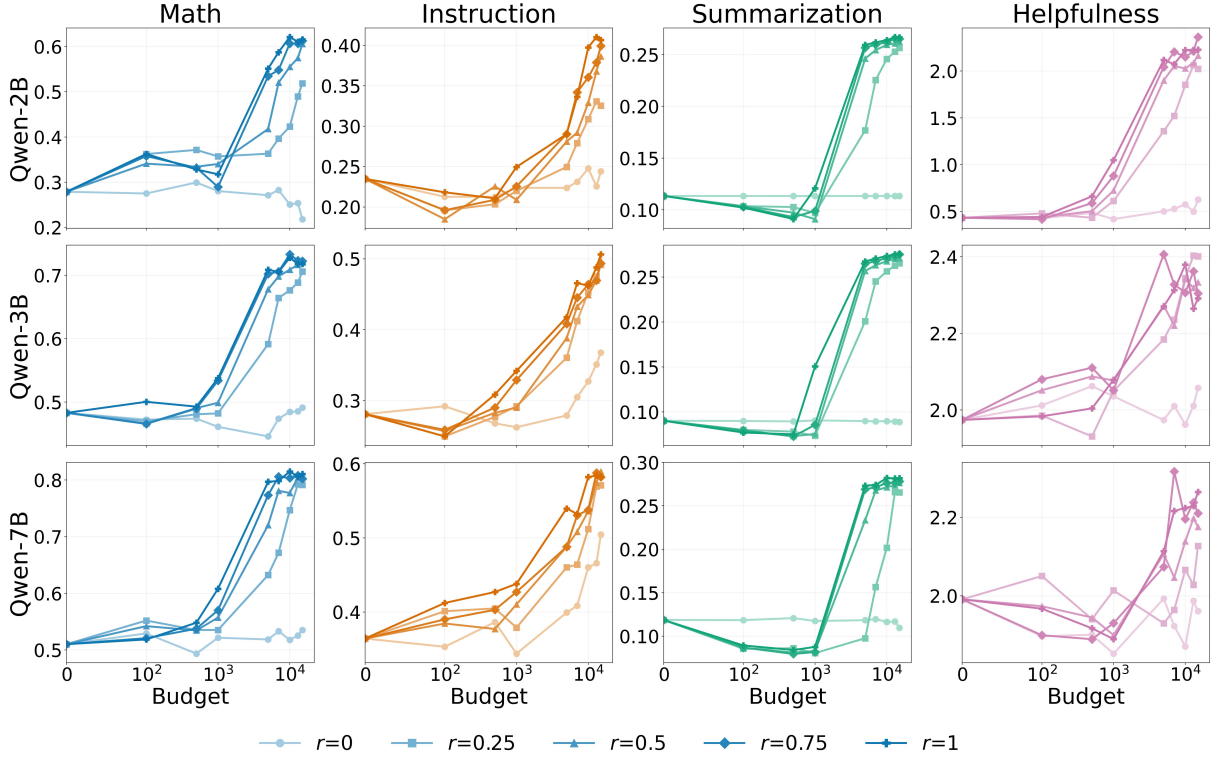}
    \caption{
    Performance versus training budget across SFT--DPO allocation ratios on the Qwen~2.5 model family. Qualitative trends align with \cref{fig:appendix-llama-3x4}. 
    }
    \label{fig:appendix-qwen-3x4}
\end{figure*}

\begin{figure*}[!ht]
    \centering
    \includegraphics[width=\textwidth]{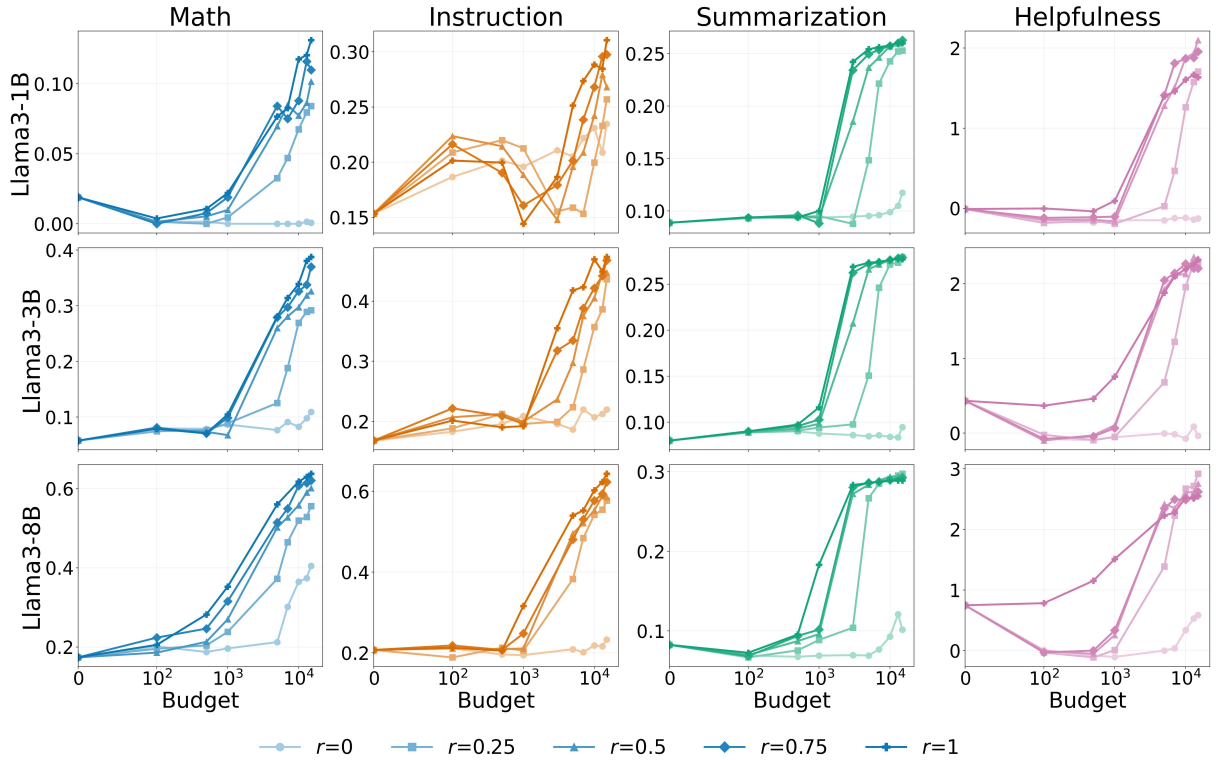}
    \caption{
    Performance versus training budget across SFT--GRPO allocation ratios on the Llama 3 model family. Qualitative trends align with \cref{fig:appendix-llama-3x4}.
    }
    \label{fig:appendix-llama-rlvr-3x4}
\end{figure*}

\begin{figure*}[!ht]
    \centering
    \includegraphics[width=\textwidth]{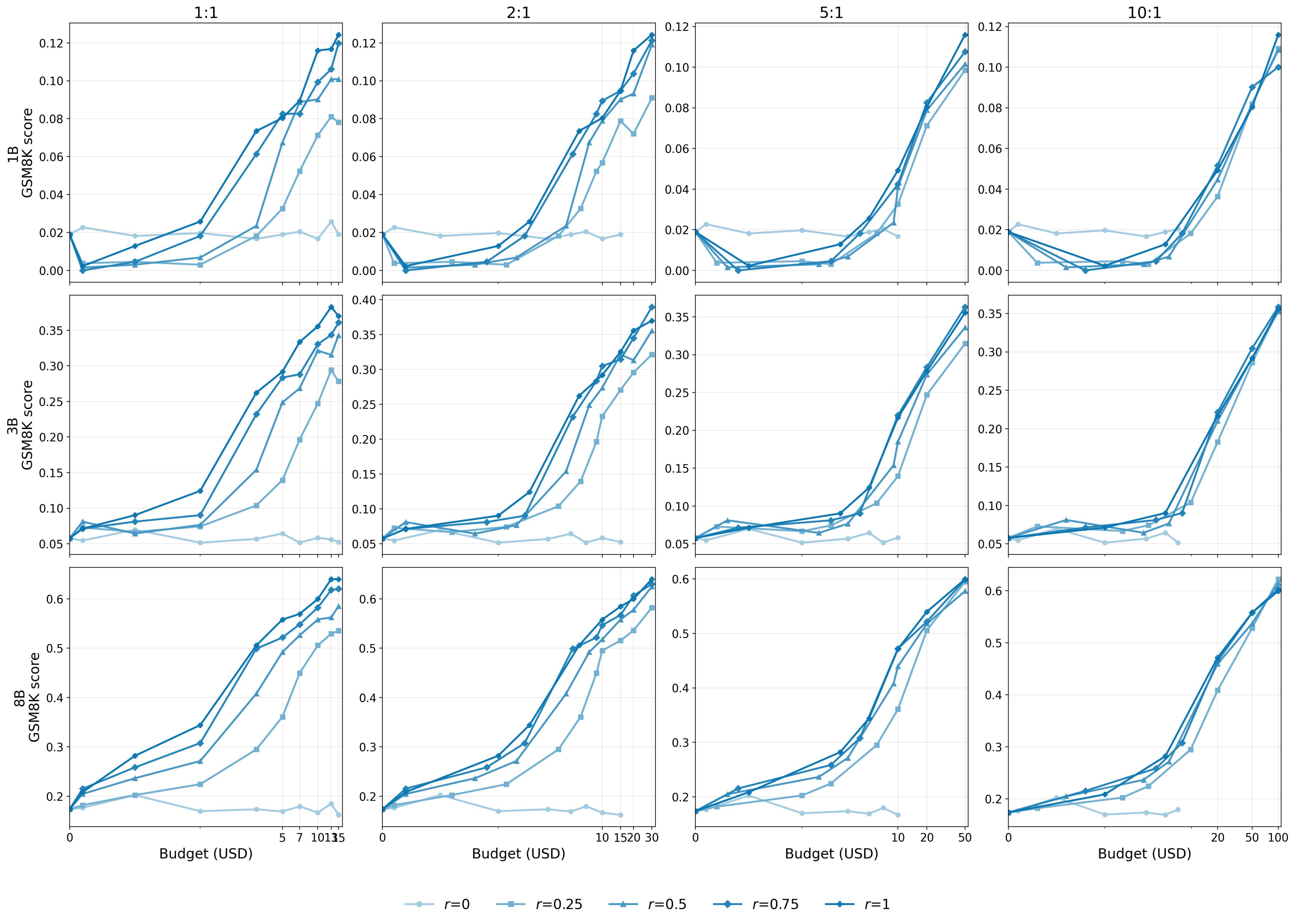}
    \caption{Full performance curves under heterogeneous SFT--DPO annotation costs for the Llama 3 model family on math task, organized by model size (rows) and SFT-to-DPO cost ratio $\rho$ in $\{1,2,5,10\}$ (columns). Within each panel, lines correspond to SFT allocation ratios $r \in \{0.0,\, 0.25,\, 0.5,\, 0.75,\, 1.0\}$; the $x$-axis shows the dollar-equivalent training budget. As $\rho$ increases (moving right within each row), the spread of curves narrows, consistent with the widening near-optimal allocation region under cost asymmetry reported in \cref{sec: money}.
    }
    \label{fig:appendix-llama-gsm8k-money-3x4}
    \vspace{-3mm}
\end{figure*}

Figs. \ref{fig:appendix-llama-3x4}--\ref{fig:appendix-llama-gsm8k-money-3x4} report raw performance curves across budgets for each allocation ratio. Two patterns are visible. 
First, in the early-stage training ($B < 5\text{k}$), especially for small models and, most notably, for instruction-following tasks in the Llama family and math tasks in the Qwen~2.5 family, the curves can be noisy and occasionally non-monotonic, so the relative ordering of allocation ratios fluctuates with budget. Because of the instability of early stages, we restrict our scaling and transfer analysis to a more stable training regime, which is consistent with established practice in scaling-behavior analysis work: power-law learning-curve fits are known to break down during early training~\citep{kaplan2020scaling}; low-compute points exhibit systematically larger fitting residuals and are routinely downweighted in scaling-law fits~\citep{hoffmann2022training}; and small-scale points can become uninformative before asymptotic regime transitions~\citep{caballero2023broken}. In the fine-tuning regime specifically, stability has been observed to require a minimum data threshold: e.g., approximately 2K examples in~\citet{zhou2023lima}. Together, these motivate an analysis of the $B \geq 5\text{k}$ regime, where post-training curves have stabilized. 
Beyond the general low-budget instability noted above, the Qwen-on-summarization panels exhibit a distinct task-specific pattern: SFT performance is non-monotonic across all model scales, first decreasing before recovering at larger budgets. This contrasts with the other (task, model) combinations, where SFT either improves monotonically or shows only a brief initial dip in the smallest model, which resolves quickly. We hypothesize that the persistent warmup in Qwen summarization may reflect a distribution mismatch. When the fine-tuning distribution differs substantially from the pretraining distribution, early SFT updates can push the model toward sub-optimal regions before sufficient task-specific signal accumulates~\citep{kotha2024understanding}. Qwen's pretraining distribution and the English-centric Reddit TL;DR distribution~\citep{stiennon2020learning} can be far in this sense. Another behavior is for Qwen 7B on helpfulness, zero-shot performance is already strong ($\sim$2.0) and post-training only marginally improves it ($\sim$2.3 at the larger budget). At the same time, the smaller Qwen 3B reaches $\sim$2.4 on the same task. A broad near-optimal region on this task for 7B may partly reflect saturation: when no allocation meaningfully improves the model, all allocations are trivially near-optimal, rather than genuine allocation invariance. 
The second pattern is that, as $B$ increases past $5\text{k}$, curves stabilize, and the gap between allocation ratios narrows visibly: multiple ratios converge to similar performance. This provides direct visual support for the broad near-optimal regions documented in \cref{sec: epsilon-region}: at larger budget, the choice of SFT--RL allocation has less impact on downstream performance, with the near-optimal region widening as budget grows. 

\section{Complete Analysis Figures with Different Settings}
\label{app:complete}

\begin{figure*}[!ht]
    \centering
    \includegraphics[width=\textwidth]{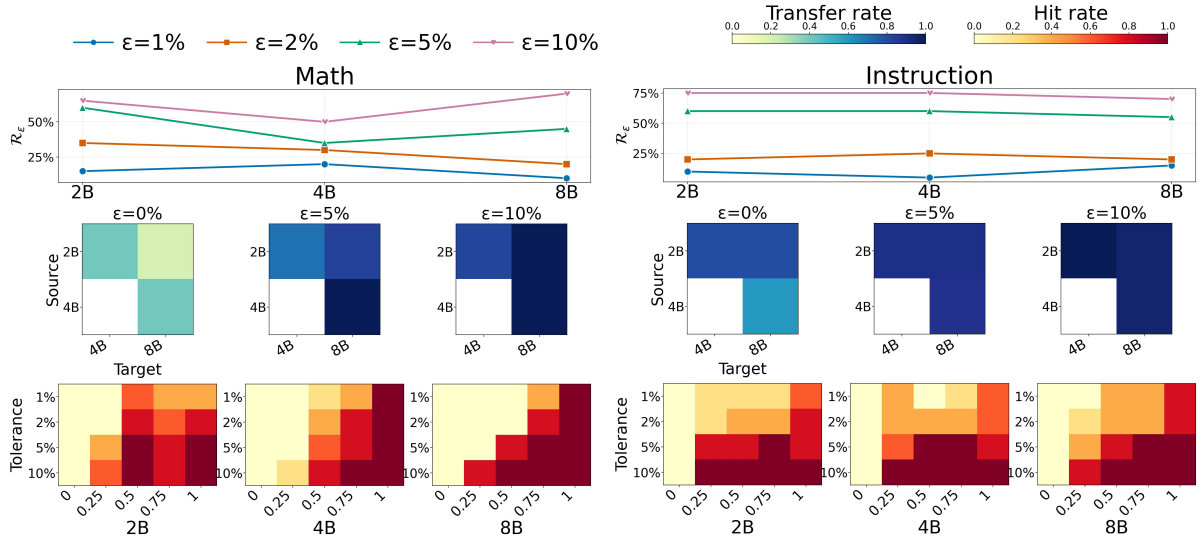}
    \caption{
    \textbf{Main analysis for Qwen3 family with SFT--DPO training stages.} For brevity, 2B in the figures denotes the Qwen3-1.7B model.
    }
    \label{fig:app-qwen3-width-complete}
\end{figure*}

\begin{figure*}[!ht]
    \centering
    \includegraphics[width=\textwidth]{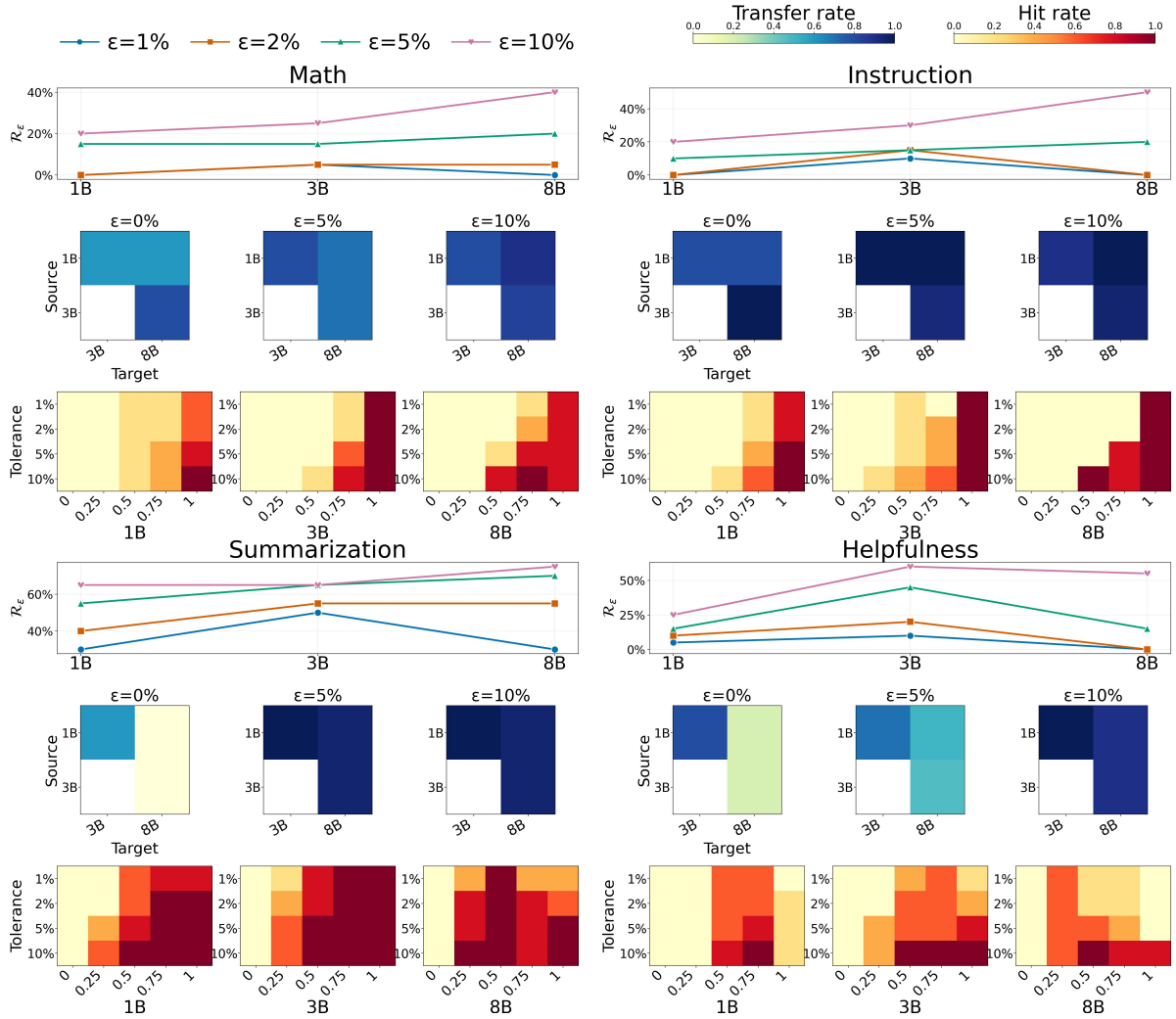}
    \caption{
    \textbf{Main analysis for Llama family with SFT--GRPO training stages.}
    }
    \label{fig:app-llama-rlvr-width-complete}
\end{figure*}

\begin{figure*}[!ht]
    \centering
    \includegraphics[width=\textwidth]{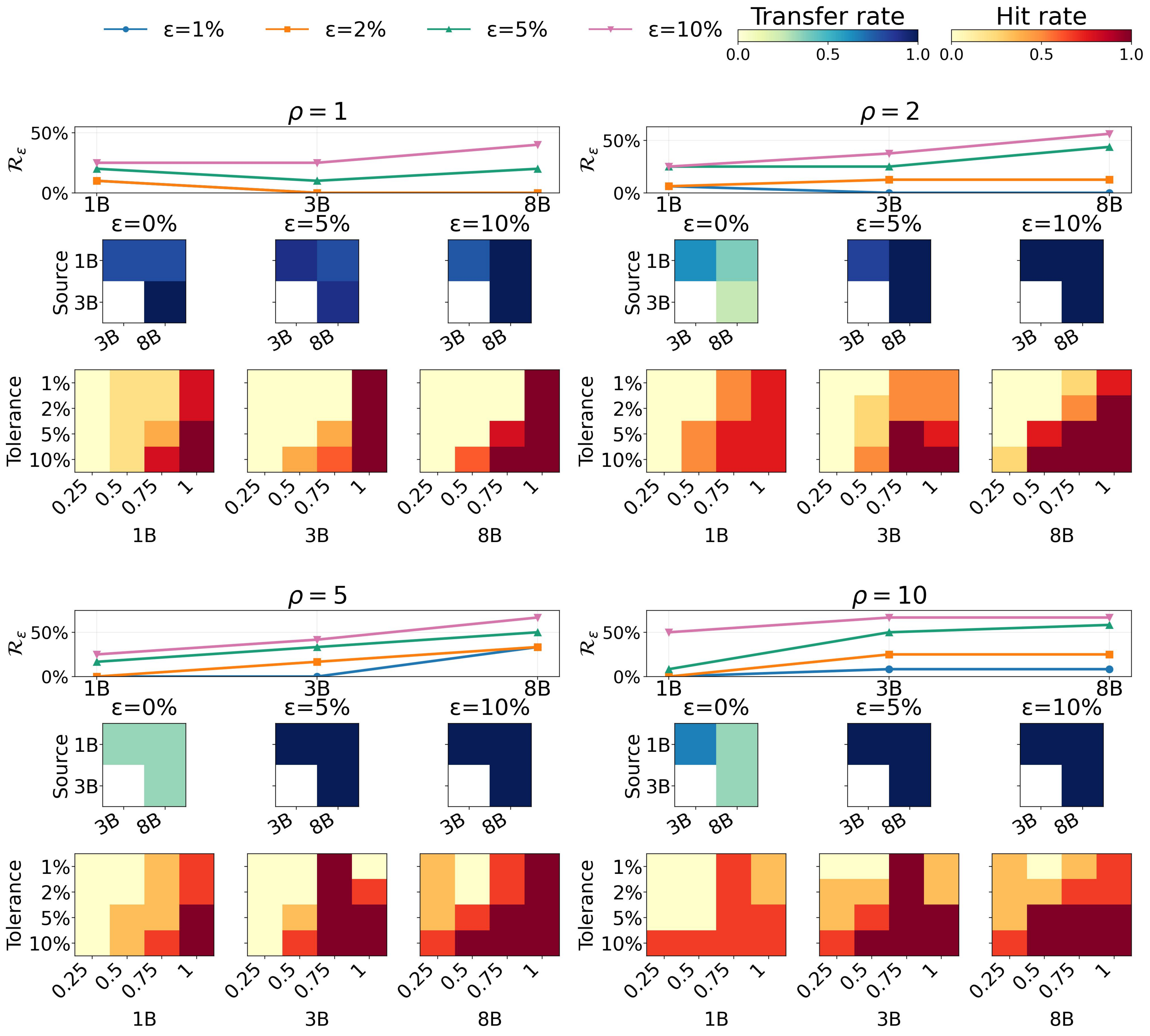}
    \caption{
    \textbf{Effect of heterogeneous annotation costs on transferable allocation regions for Math task.}
    }
    \label{fig:app-llama-gsm8k-width-complete}
\end{figure*}

We list the complete analysis figures with mean tolerance-optimal region vs. model size in the first row; cross-size overlap of tolerance-optimal ratio regions in the second row, and per-ratio hit rate across budgets in the third row for Qwen~3 family in \cref{fig:app-qwen3-width-complete}, SFT--GRPO setting in \cref{fig:app-llama-rlvr-width-complete}, and cost asymmetry setting in \cref{fig:app-llama-gsm8k-width-complete}. 

\section{Analysis with Additional Model Family and Model Scale}
\label{app:extended-scale}

\begin{figure*}[!ht]
    \centering
    \includegraphics[width=\textwidth]{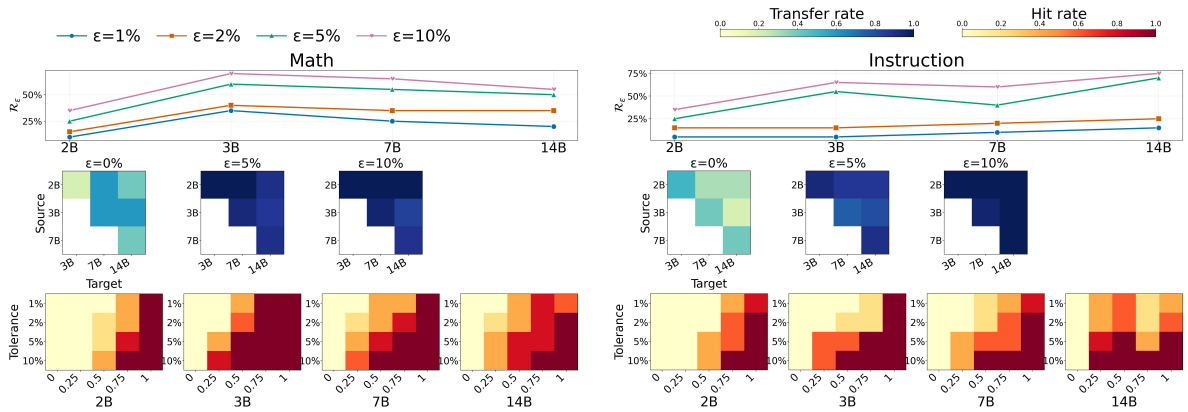}
    \caption{
    \textbf{Main analysis for Qwen2.5 family up to 14B model scale with SFT--DPO training stages.} 
    }
    \label{fig:app-qwen14-width-full}
\end{figure*}

To test whether these trends are tied to the scales and families of our main
sweep, we extend the analysis along two axes: a more recent family (Qwen~3, 1.7B/4B/8B), and a larger target within Qwen~2.5
(14B), on math and instruction
following, with results in \cref{fig:app-qwen3-width-complete} and \cref{fig:app-qwen14-width-full}. 

While the near-optimal region generally trends wider with scale, we observe that this widening is occasionally non-monotonic, with slight contractions at certain scales. However, the 14B scale is far wider than the 2B (Qwen2.5 1.5B model) proxy; on math, the width can dip slightly at 14B while transfer stays strong.
Crucially, our transferability remains highly robust despite these local width fluctuations, further validating the reliability of the near-optimal region.

\section{Additional Ablation Study}
\subsection{Ablation Study 1: Grid Resolution}
\label{app:dense-grid}

\begin{figure*}[!ht]
    \centering
    \includegraphics[width=\textwidth]{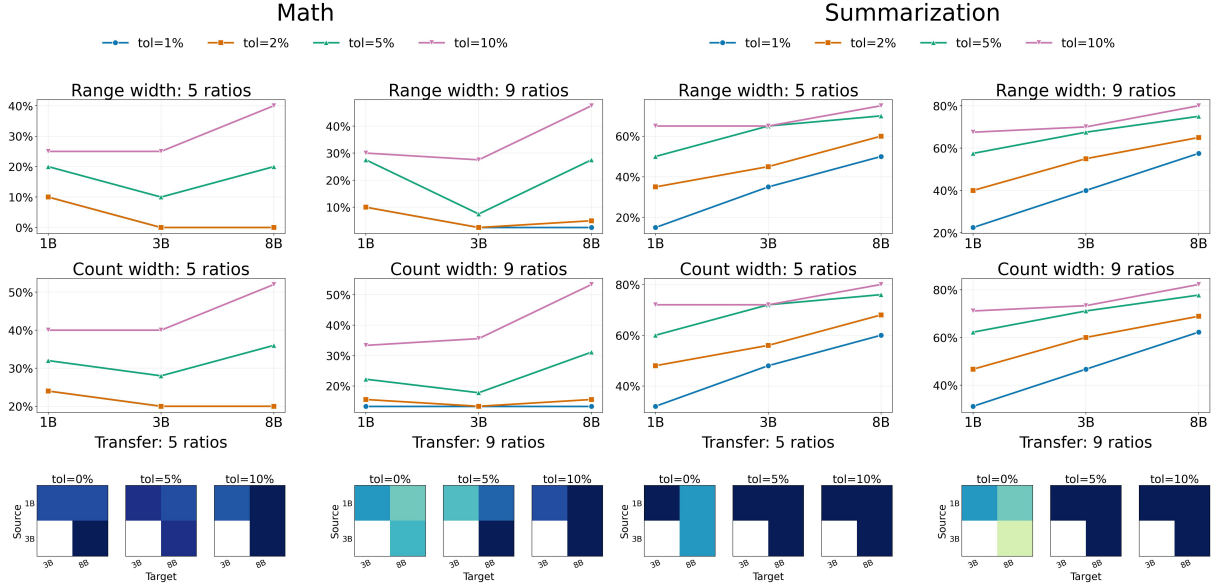}
    \caption{Validation of the 5-point allocation grid on math and summarization task. For each task: Left: range-width $w_\epsilon^{\,\mathrm{rng}}$; $w_\epsilon^{\,\mathrm{cnt}}$ and transfer hit computed from the 5-point grid $\mathcal{G} = \{0.00, 0.25, 0.50, 0.75, 1.00\}$. Right: range-width computed from a denser 9-point grid adding intermediate ratios $\{0.125, 0.375, 0.625, 0.875\}$. Both grids yield qualitatively similar trends across tolerance and model scale, supporting the use of the 5-point grid as the primary protocol. }
    \label{fig:dense-grid}
\end{figure*}

To assess whether the 5-point allocation grid $\mathcal{G} =\{0.00, 0.25, 0.50, 0.75, 1.00\}$ used in our main experiments is sufficient for the conclusions we draw, we conduct a density-validation study on math (GSM8K) and the summarization task for the Llama 3 family. We compare the range-width estimator $w_\epsilon^{\,\mathrm{rng}}$ and count-width estimator $w_\epsilon^{\,\mathrm{cnt}}$ computed from the 5-point grid against the same estimators computed from a denser 9-point grid that adds intermediate ratios
$\{0.125, 0.375, 0.625, 0.875\}$:
\[
\begin{aligned}
\mathcal{G}_9
= \{&0.000, 0.125, 0.250, 0.375, 0.500,\\
   &0.625, 0.750, 0.875, 1.000\}.
\end{aligned}
\]
  
\cref{fig:dense-grid} reports the estimates from both grids across tolerance thresholds and model scales. Near-optimal range width and transferability trends are qualitatively preserved. The 9-point estimates agree with the 5-point estimates in direction and approximate magnitude across tested $(N, \epsilon)$ combinations, supporting the use of the 5-point grid as our protocol.
The results confirm the 5-point grid resolution is sufficient to capture the qualitative scaling and transferability behavior reported in the main text at a tractable experimental cost.

\subsection{Ablation Study 2: Count-Width Analysis}
\label{app:count}

\begin{figure*}[!ht]
    \centering
    \includegraphics[width=0.5\textwidth]{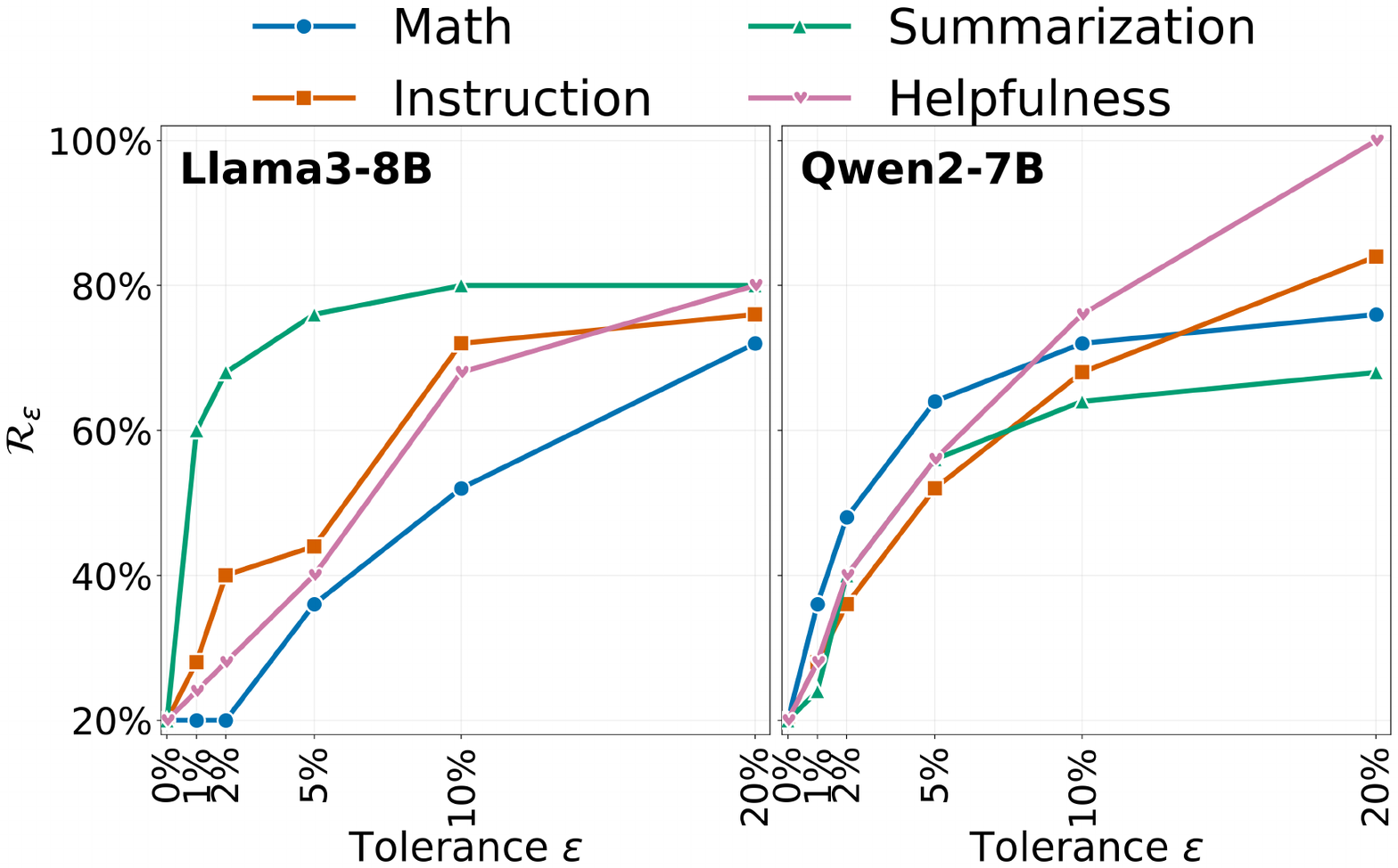}
    \caption{Tolerance $\epsilon$ versus near-optimal-region width $\mathcal{R}_\epsilon = (w_{\epsilon}^{\text{cnt}})$ for Llama3-8B and Qwen2.5-7B across four tasks. The near-optimal region expands consistently with tolerance across tasks and model families.}
    \label{fig:appendix-tolerance-count}
\end{figure*}

\begin{figure*}[!ht]
    \centering
    \includegraphics[width=\textwidth]{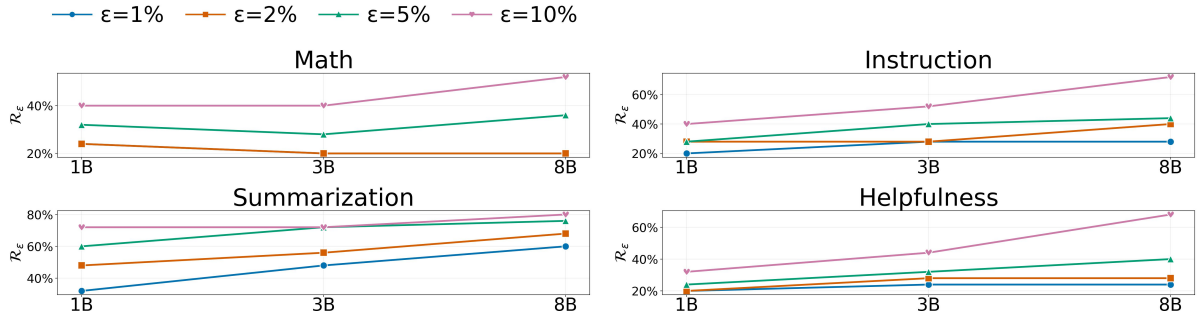}
    \caption{Near-optimal-region width $\mathcal{R}_\epsilon = (w_{\epsilon}^{\text{cnt}})$ vs. model size $N$ for Llama-family. Each panel shows one task; lines correspond to different tolerance thresholds $\epsilon$. Trends qualitatively replicate the range-width results in \cref{fig:main-llama-width}. The corresponding transfer rates and per-ratio hit-rate heatmaps are estimator-invariant and reported in the main paper.}
    \label{fig:appendix-llama-count}
\end{figure*}

\begin{figure*}[!ht]
    \centering
    \includegraphics[width=\textwidth]{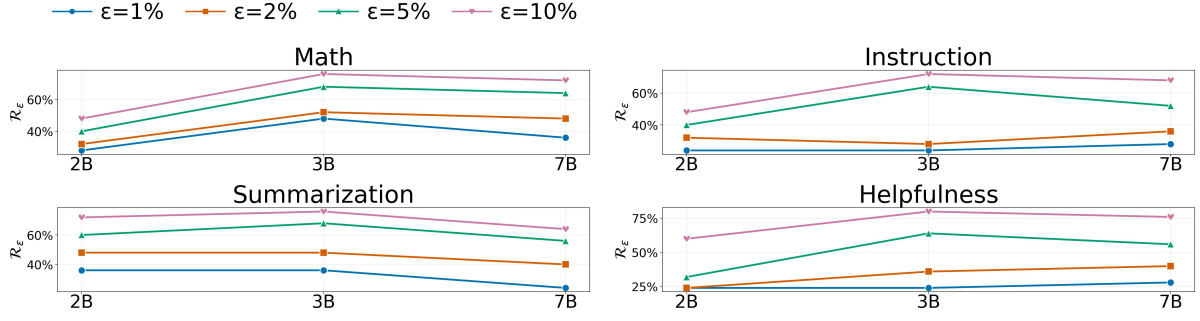}
    \caption{
    Near-optimal-region width $\mathcal{R}_\epsilon = (w_{\epsilon}^{\text{cnt}})$ vs. model size $N$ for Qwen~2.5-family.
    }
    \label{fig:appendix-qwen-count}
\end{figure*}

\begin{figure*}[!ht]
    \centering
    \includegraphics[width=\textwidth]{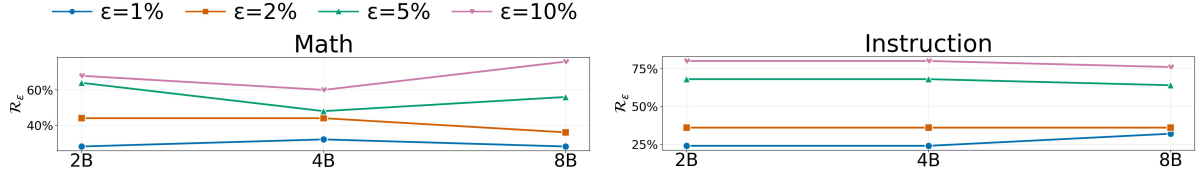}
    \caption{
    Near-optimal-region width $\mathcal{R}_\epsilon = (w_{\epsilon}^{\text{cnt}})$ vs. model size $N$ for Qwen~3-family.
    }
    \label{fig:appendix-qwen3-count}
\end{figure*}

\begin{figure*}[!ht]
    \centering
    \includegraphics[width=\textwidth]{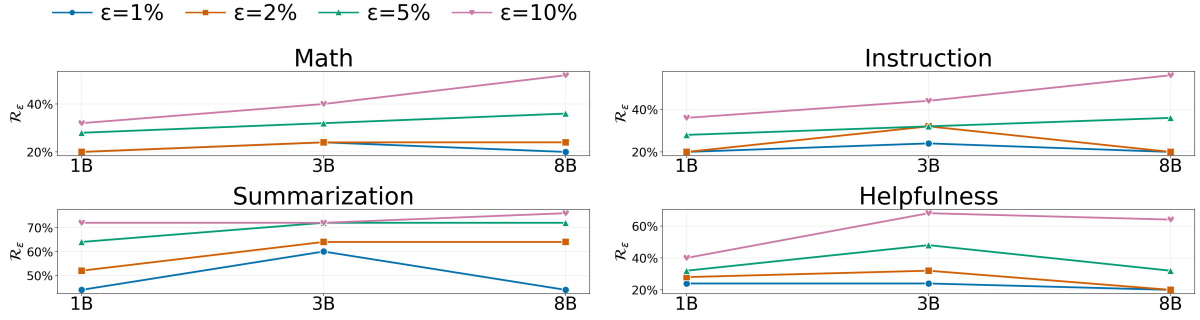}
    \caption{
    Near-optimal-region width $\mathcal{R}_\epsilon = (w_{\epsilon}^{\text{cnt}})$ vs. model size $N$ for Llama-family for SFT:GRPO Setting.
    }
    \label{fig:appendix-llama-rlvr-count}
\end{figure*}

\begin{figure*}[!ht]
    \centering
    \includegraphics[width=0.7\textwidth]{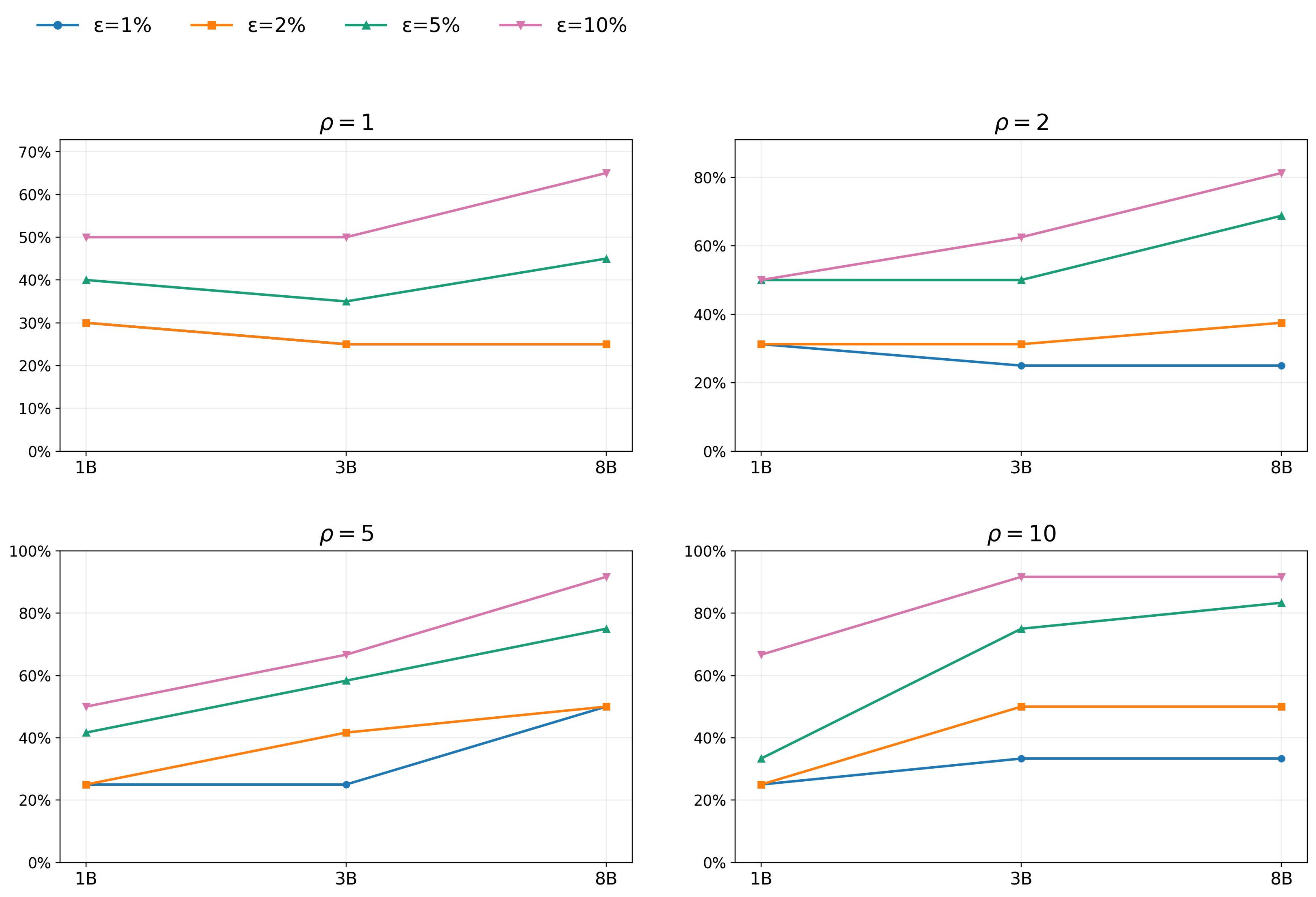}
    \caption{
    Near-optimal-region width $\mathcal{R}_\epsilon = (w_{\epsilon}^{\text{cnt}})$ vs. model size $N$ for Llama-family for cost-asymmetric setting.
    }
    \label{fig:appendix-money-gsm8k-count}
    \vspace{-3mm}
\end{figure*}

We complement the range-width analysis in the main text with the count-width estimator
\[
w_\epsilon^{\,\mathrm{cnt}}(N, B) \;=\;
|\widehat{\mathcal{R}}_\epsilon(N, B)|\,/\,|\mathcal{G}| \ ,
\]
the fraction of evaluated grid candidates within the $\epsilon$-near-optimal region (\cref{sec:post-training-allocation}). Unlike the range estimator, count is sensitive to grid gaps and therefore directly diagnoses whether $\widehat{\mathcal{R}}_\epsilon$ is non-contiguous on the grid. The choice of range vs. count affects only the scalar width statistic (Row~1 of each main-paper analysis figure); the transfer rate $T_\epsilon$ and per-ratio hit-rate heatmaps (Rows~2 and~3) depend only on the set $\widehat{\mathcal{R}}_\epsilon$ itself and are therefore identical to those reported in the main paper regardless of width estimator. We accordingly report only the count-width vs. model-size panels here and refer the reader to the main text and \cref{app:complete} for transfer and hit-rate results.

Figs.~\ref{fig:appendix-tolerance-count}--\ref{fig:appendix-money-gsm8k-count} report the count-width analysis across six experimental settings: sensitivity of budget allocation (\cref{fig:tolerance-width}), Llama SFT--DPO  (\cref{fig:main-llama-width}), Qwen~2.5 SFT--DPO (\cref{fig:main-qwen-width}), Qwen~3 SFT--DPO (\cref{fig:main-qwen3-width}), Llama SFT--GRPO (\cref{fig:llama-rlvr-width}), and the heterogeneous-cost setting under DPO (\cref{fig: cost_math}). Across all settings, the count-based results generally replicate the qualitative trends of the range-width analysis in the main text: (i) the near-optimal region widens with tolerance at all model scales; and (ii) width increases with model scale under a fixed tolerance.

The qualitative agreement between range and count estimators supports the contiguity argument from \cref{sec:post-training-allocation}.
The observed agreement, along with the previous ablation study (\cref{app:dense-grid}),  empirically validates treating $\widehat{\mathcal{R}}_\epsilon$ as approximately contiguous on $\mathcal{G}$ and justifies our choice to report range-based width as the primary statistic in the main text.

\subsection{Ablation Study 3: Multi-Seed Validation}
\label{app:seed-ablation} 
\begin{figure*}[!ht]
    \centering
    \includegraphics[width=\textwidth]{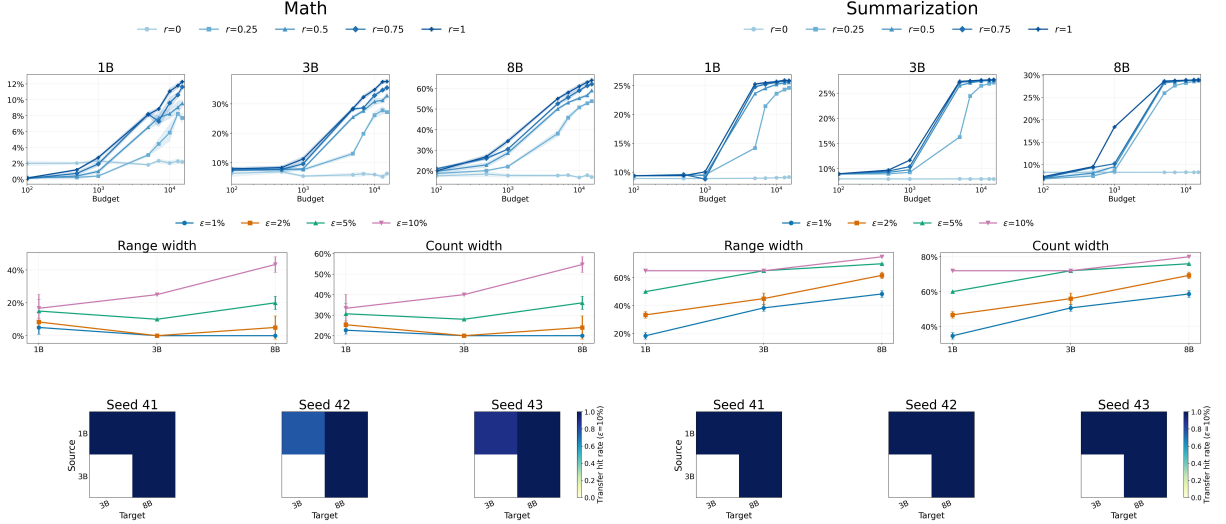}
    \caption{Three-seed validation on math (left) and summarization (right) (Llama 3 family). \textbf{Row 1:} Performance curves versus training budget across SFT--DPO allocation ratios; lines show mean and shaded bands show $\pm 1$ standard deviation across 3 seeds. \textbf{Row 2:} Mean near-optimal region width versus model size $N$ per tolerance level $\epsilon$; lines show mean and bands show $\pm 1$ standard deviation across seeds. \textbf{Row 3:} Per-seed cross-scale transfer hit-rate matrices at $\epsilon = 10\%$; each column corresponds to one seed. Seed variation perturbs both the choice of training samples and the training order. Mean trends in Rows 1--2 are stable across seeds, and the qualitative transfer pattern in Row 3 is consistent across all three seeds, supporting the validity of qualitative analyses discovered with single-seed estimates.}
    \label{fig:multi-seed}
\end{figure*}

The full experimental sweep covers the Cartesian product of model sizes, allocation ratios, budgets, tasks, algorithms (DPO, GRPO), and model families (Llama 3, Qwen 2.5). Running each configuration with multiple seeds would multiply this cost by the number of seeds. We therefore report single-seed point estimates in the main text and conduct a 3-seed validation to assess whether the single-seed estimates are sufficient for the qualitative trends we report. We run this validation on math (GSM8K) and summarization for the Llama 3 family, two of the more compute-intensive settings in our sweep. Seed variation perturbs both the random subset of training samples drawn from the source pool and the training order within each stage.

\cref{fig:multi-seed} reports three views:
\begin{itemize}
  \itemsep1pt
  \item \textbf{Row 1 (performance curves):} mean performance versus budget across SFT--DPO allocation ratios is generally stable across seeds.
  \item \textbf{Row 2 (near-optimal region width versus tolerance):} the mean width curve is generally consistent across seeds in shape. 
  \item \textbf{Row 3 (cross-scale transfer hit rates at $\epsilon = 10\%$):} We report per-seed hit-rate matrices rather than averaging, since transfer hit rates are integer-valued and averaging across seeds yields fractional values that are relatively hard to interpret. The qualitative trend, higher hit rates at larger tolerance, similar magnitude across all three seeds, is preserved.
\end{itemize}

Together, these results support the use of single-seed point estimates for the qualitative phenomena reported in the main text. Obtaining multi-seed variance bars across the full experimental sweep would require substantially more compute. However, the consistency observed here suggests that quantitative variance bars would not change the qualitative conclusions.

\subsection{Ablation Study 4: Low-Budget Regime}
\label{app:low-budget}

\begin{figure*}[!ht]
    \centering
    \includegraphics[width=\textwidth]{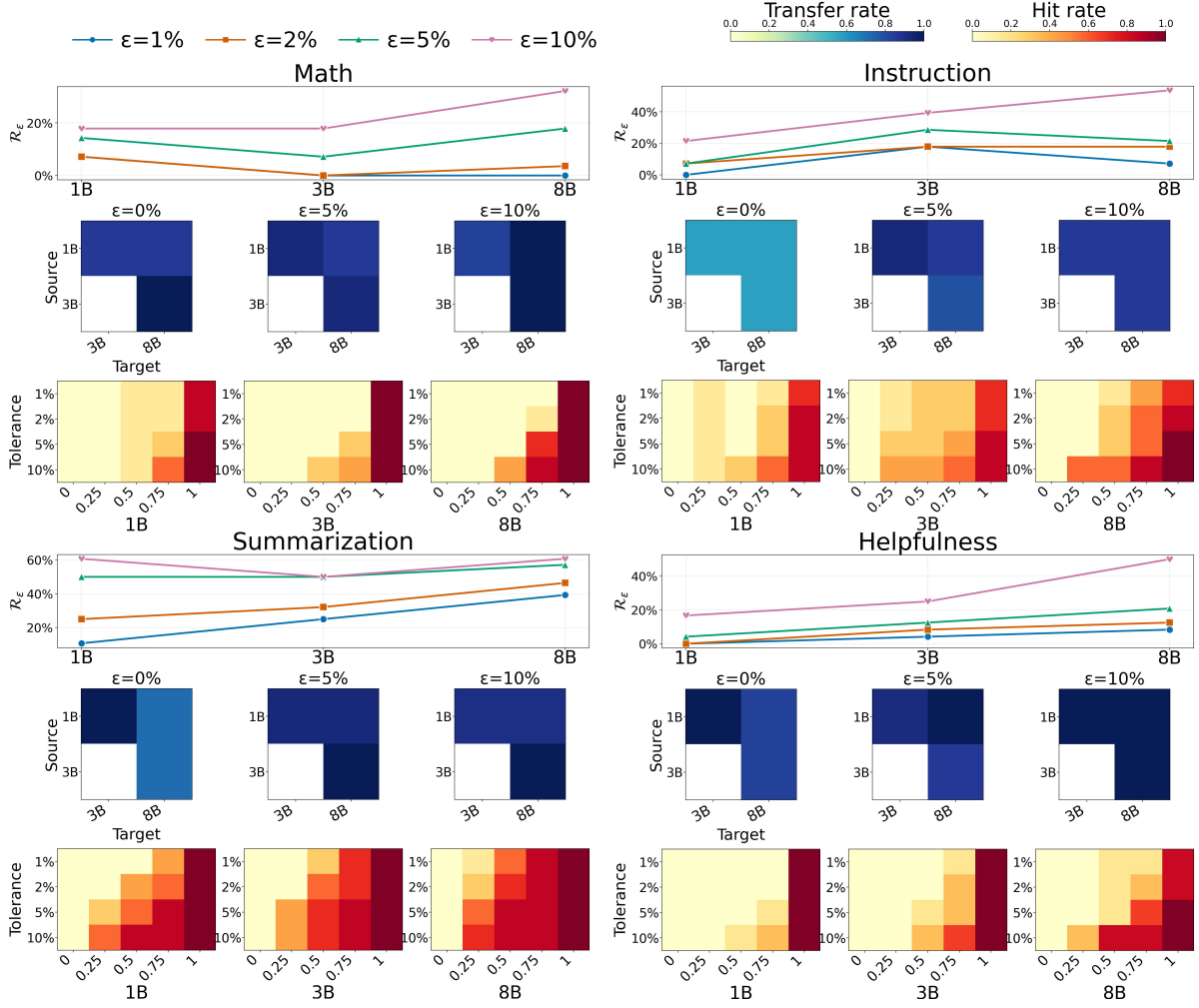}
    \caption{Main analysis for Llama family with SFT--DPO training stages with $B \geq 1\text{k}$. }
    \label{fig:low-budget-ablation}
\end{figure*}

As discussed in \cref{sec:post-training-allocation} and \cref{app:curves}, performance curves in the small-budget regime can be noisy and occasionally non-monotonic. The main analysis, therefore, focuses on the stable $B \geq 5\text{k}$ regime where cross-model-scale scaling and transferability trends can be cleanly isolated. This ablation extends the analysis to lower budgets to verify that the qualitative findings persist, though somewhat attenuated when the unstable low-data regime is included. 
\cref{fig:low-budget-ablation} reports the extended-range results. Here are our two key observations:
\begin{itemize}
  \itemsep1pt
  \item \textbf{Qualitative trends persist when low-budget points are included.} The widening of $\mathcal{R}_\epsilon$ with model scale and the cross-scale transferability of near-optimal regions remain observable even when budgets below $5\text{k}$ are included, confirming that the main-text findings are not contingent on the $B \geq 5\text{k}$ regime.
  
  \item \textbf{Attenuated effects at the lower budgets, especially for IFEval.} Consistent with the run-to-run instability documented in \cref{app:curves}, the cross-scale trends in $\mathcal{R}_\epsilon$ and $T_\epsilon$ are weaker when smaller budgets are included, most noticeably for instruction-following (IFEval). This attenuation reflects the fact that noise can dominate the signal at a lower budget, rather than a breakdown of the underlying transferability mechanism. 
\end{itemize}

At $B \sim 1\text{k}$, the post-training pipeline can be exhaustively swept directly at lower cost, and the optimal allocation shifts toward SFT-only configurations~\citep{raghavendra2025balancing}. Our transferability framework is not intended to replace direct grid search in this regime; our value emerges when full-scale search becomes computationally prohibitive: the regime of larger budgets, past the early-training phase, that we report on most directly.

\subsection{Ablation Study 5: Robustness to Training Hyperparameters}

\begin{figure*}[!ht]
    \centering
    \includegraphics[width=\textwidth]{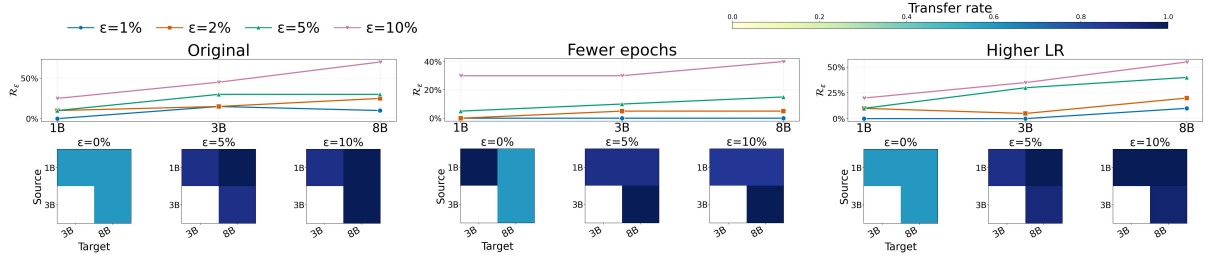}
    \caption{
    Changing training configurations on the instruction-following task for the Llama family.
    }
    \label{fig:appendix-llama-ifeval-hyperparam}
\end{figure*}

Our main sweep fixes one SFT and one RL configuration per family. To check that the trends are not an artifact of that choice, we re-ran the instruction-following analysis on the Llama family (1B/3B/8B) under two perturbations having fewer epochs or doubled the learning rate during training, and recomputed width and transfer, with results in \cref{fig:appendix-llama-ifeval-hyperparam}.
The two qualitative trends persist in both cases: width is non-decreasing in scale at every tolerance and strictly increasing at $\varepsilon = 5\%$ and $10\%$ (it is identically $0.00$ at $\varepsilon = 1\%$ with fewer epochs), and near-optimal transfer exceeds exact-optimum transfer.

\subsection{Ablation Study 6: Absolute-Tolerance Ratio}
\label{app:abs-tolerance}

\begin{figure*}[!ht]
    \centering
    \begin{subfigure}[t]{\textwidth}
        \centering
        \includegraphics[width=\textwidth]{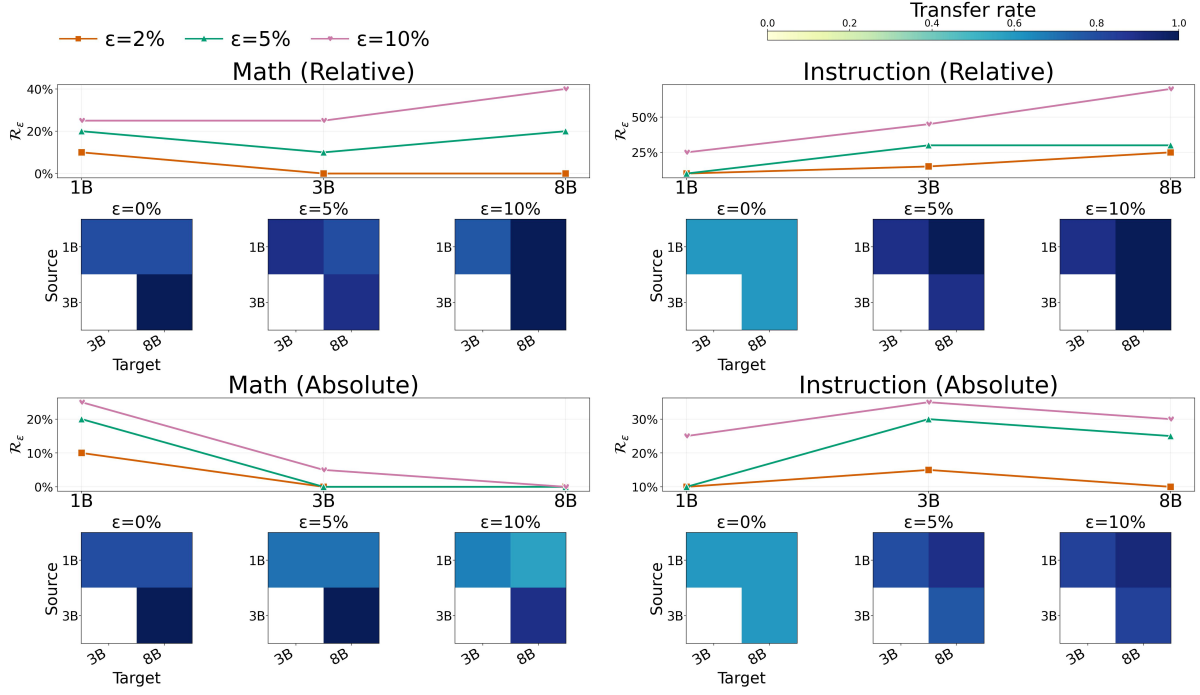}
        \caption{Llama Model Family.}
        \label{fig:appendix-llama-fixed}
    \end{subfigure}
    \\
    \begin{subfigure}[t]{\textwidth}
        \centering
        \includegraphics[width=\textwidth]{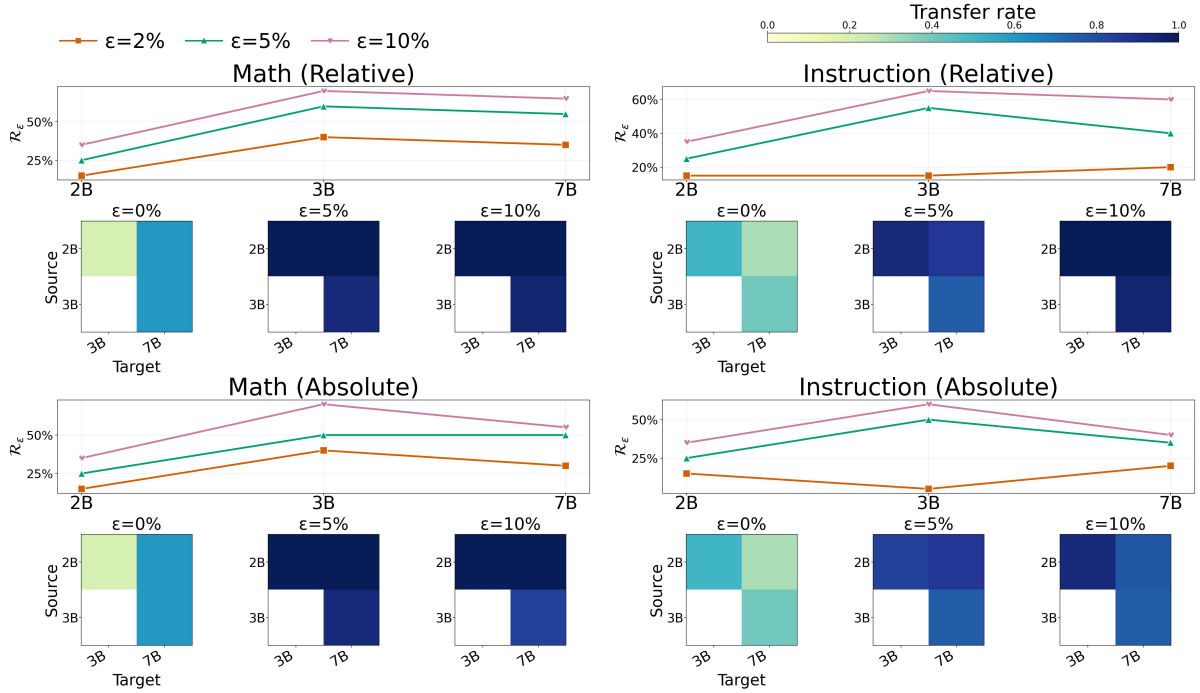}
        \caption{Qwen 2.5 Model Family.}
        \label{fig:appendix-qwen-fixed}
    \end{subfigure}
    \caption{Main analysis results under a relative tolerance (main paper) versus an absolute tolerance. }
    \label{fig:appendix-fixed-ratio}
\end{figure*}

A relative tolerance can admit more ratios simply because larger models have higher peak performance. To test this directly, we re-ran our analysis under an \textit{absolute} tolerance (a fixed performance margin anchored to the smallest model, as an arbitrary margin range does not make sense). Specifically: Let $N_{\min} = \min \cN$ denote the smallest model size (in our setting, 1B for Llama, 1.5B for Qwen2.5 (denoted by 2B)). 
We define an absolute performance tolerance, anchored to the smallest model, as
$
    \Delta_\epsilon(B)
    := \epsilon \max_{r' \in \cG} \cP(N_{\min}, r', B).
$
We then estimate the near-optimal region by
\als{
    \widehat{\cR}^{\mathrm{abs}}_\epsilon(N, B)
    &= \Big\{\, r \in \cG:
    \cP(N,r,B) \\
    &\quad \geq
    \max_{r' \in \cG} \cP(N,r',B)
    - \Delta_\epsilon(B)
    \,\Big\} \ .
}
The results comparing $\widehat{\cR}_\epsilon(N, B)$ (top row, reports Range Width) and $\widehat{\cR}^{\mathrm{abs}}_\epsilon(N, B)$ (bottom row, reports Range Width) are in \cref{fig:appendix-fixed-ratio}.

\paragraph{1. Width: the widening is not merely a relative-tolerance artifact.} An absolute tolerance largely preserves the scaling behavior seen under a relative tolerance for the Qwen family. For the Llama family, the widening is dampened (e.g., Llama-Inst widens continuously under a $10\%$ relative tolerance, but increases and then decreases under an absolute tolerance). The starkest contrast is Llama-Math, where the absolute slope is negative. This contrast between Qwen-math and Llama-math's absolute tolerance behavior is due to Llama's small model performing far below its 8B counterpart, so a fixed absolute margin spans a very different share of the performance range at each scale. This is precisely why \textit{an absolute tolerance is task- and model-dependent} (\cref{sec: scaling-law}), and why we adopt the practitioner-relevant relative criterion: it answers ``how much allocation freedom retains $X\%$ of \textit{this} model's peak?'' without anchoring to a family whose behavior differs sharply across scale.

\paragraph{2. Transfer: Robust under both tolerances.}

Near-optimal regions transfer far better than the exact optimum in 3 out of 4 settings. This addresses the possibility that transfer is driven by mechanical overlap: under an absolute tolerance, the target region is generally \textit{narrower} than under a relative one (e.g., Qwen-Inst 7B), yet transfer stays high. Mechanical overlap from a wider target would predict the opposite. The exception is  Llama-Math, where the 8B absolute region remains a single ratio (width $0.00$) even as the absolute tolerance increases because the fixed performance margin anchored to the smallest model is way too small for 8B. Here, the exact optimum transfers unusually well precisely because that single ratio is stable across scale, so ``near-optimal vs. exact'' is degenerate in this setting rather than a genuine reversal of our claim. 

To summarize, peak-performance scaling does marginally contribute to the widening effect (visible in the Llama-math results), but our core practical claim, transferability, remains generally robust under both tolerance definitions. Where absolute transfer drops (Llama-Math), it reflects the task- and model-dependent nature of an absolute margin, which is itself the reason a relative tolerance is a more appropriate metric.

\subsection{Ablation Study 7: Proxy Versus Fixed Ratio Comparison}

We observe that a $10\%$ tolerance makes the region wide at scale, so intuitively, a centered default often lands within it. However, a fixed $r=0.5$ breaks in two common cases: (i) a stricter tolerance (e.g., $5\%$, retaining $95\%$ of peak); and (ii) an off-center region, where the optimum does not sit near $0.5$. The proxy handles both, because it transfers the region's \textit{location}, which we show is reliable across scales. We also ran an experiment for a small proxy 5-point search vs. a fixed $r=0.5$, and report reliability and cost below. We measure how often each recommendation lands within the target model's near-optimal region (proxy: ratios transferred from the 1B/1.5B models), averaged over budgets and target sizes (3B–8B). As shown in \cref{tab:app_transferability_comparison}, at both tolerances, the proxy generally outperforms or is comparable to a fixed $r=0.5$, but this is more evident at the smaller tolerance ($0.90\text{--}0.95$ vs. $0.20\text{--}0.60$). Crucially, whether $r=0.5$ suffices depends on the task, model family, and the statistics we cannot know in advance.

Additionally, using our measured runtimes and CoreWeave L40 pricing (\$1.25/GPU-hr), a 5-point ratio search on the 1B proxy (e.g., $10\text{k}$/run) and a single 8B run at one ratio ($10k$) take the same $\sim100$ GPU-minutes $\sim$ \$2.1. A blind guess of $r=0.5$ costs the same $100$ minutes but, per \cref{tab:app_transferability_comparison}, carries real risk: it misses the $5\%$ near-optimal region up to $80\%$ of the time (e.g., Llama HelpSteer). Spending the extra $100$ minutes up front on the 1B proxy serves as an efficient `insurance policy,' providing a $> 90\%$ guarantee of hitting the target's 5\% near-optimal region on the first try. Moreover, the proxy approach (proxy + 8B run $\sim200$ GPU-minutes) is $2.5\times$ cheaper than a full 5-point sweep on the 8B target ($\sim500$ GPU-minutes). The proxy method thus offers a good trade-off between computational cost and guaranteed performance.

\begin{table}[!ht]
\centering
\caption{Transferability comparison: small model proxy vs. fixed $r=0.5$. We select these settings to vary one axis at a time: instruction-following on both Llama and Qwen~2.5 (same task, different family) and HelpSteer vs. instruction-following on Llama (same family, different task), so the comparison covers both task- and family-level.}
\label{tab:app_transferability_comparison}

\small
\setlength{\tabcolsep}{3.5pt}
\begin{tabular}{lccc}
\toprule
Task & Tol. & Proxy & Fixed $r{=}0.5$ \\
\midrule
Llama HelpSteer & $5\%$  & $0.95$ & $0.20$ \\
Llama Inst      & $5\%$  & $0.95$ & $0.40$ \\
Qwen~2.5 Inst       & $5\%$  & $0.90$ & $0.60$ \\
Llama HelpSteer & $10\%$ & $1.00$ & $0.70$ \\
Llama Inst      & $10\%$ & $0.95$ & $0.70$ \\
Qwen~2.5 Inst       & $10\%$ & $1.00$ & $1.00$ \\
\bottomrule
\end{tabular}
\end{table}

\section{Additional Results for Cost Asymmetry Setting}
\label{app:cost}
\begin{figure*}[!ht]
    \centering
    \includegraphics[width=\textwidth]{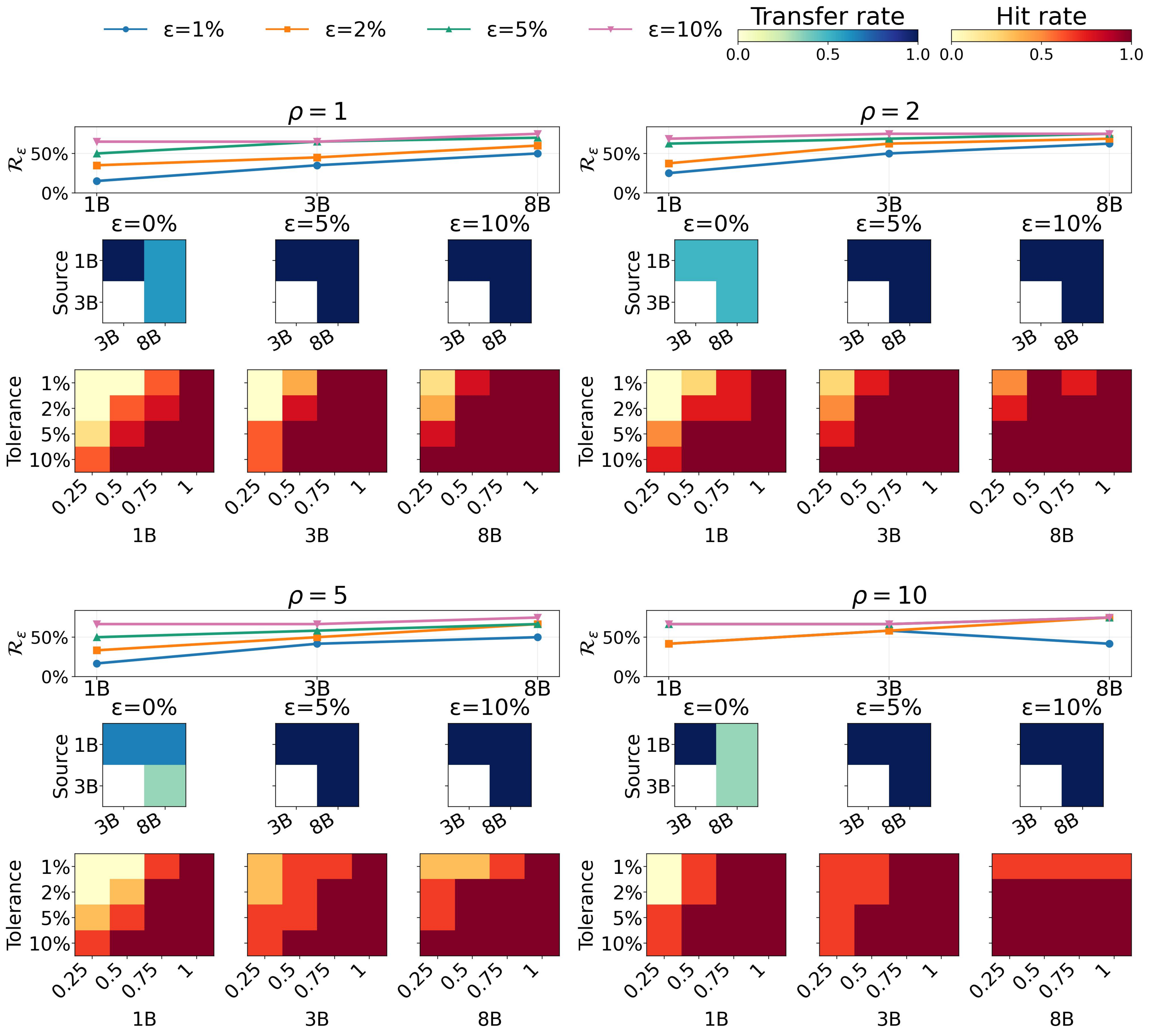}
    \caption{
    Heterogeneous annotation costs settings for the Summarization task.
    }
    \label{fig:appendix-summarization-money}
\end{figure*}

We extend the heterogeneous-cost analysis from \cref{sec: money} to the Summarization task. The experimental setup follows the same experimental protocol.
\cref{fig:appendix-summarization-money} reports the results. The qualitative scaling behavior matches the math task results in the main text: across cost settings, larger tolerance thresholds continue to produce broader $\mathcal{R}_\epsilon$ and stronger cross-scale transfer $T_\epsilon$; as $\rho$ increases, $\mathcal{R}_\epsilon$ generally widens at the same tolerance, indicating increased allocation flexibility for practitioners when SFT becomes relatively expensive. These results confirm that the cost-asymmetry findings in \cref{sec: money} are not specific to mathematical reasoning and can generalize across our task suite.

Additionally, we expand the analysis to a GPU-time budget case study. If a practitioner defines the budget purely by GPU compute time (proportional to training FLOPs~\citep{hoffmann2022training}), our framework still applies. Since DPO takes $\sim2\text{--}3\times$ the GPU time of SFT, a compute-time budget is an instance of our cost-asymmetry analysis (\cref{sec: money}) with $\rho= c_{\mathrm{SFT}}/c_{\mathrm{DPO}} \approx 0.5$ $(1{:}2)$, the regime where the RL stage is the more expensive one, extending the $\rho \geq 1$ range of the main paper. We ran this setting on math (Llama). At $\rho=0.5$, the near-optimal region reduces to essentially a single ratio at small scale (\cref{tab:app_near_optimal_width}, width $0$), and that ratio is identical across model sizes. Hence, the proxy's recommendation is always valid on the larger target, giving consistently good transfer (\cref{tab:app_cross_scale_transfer}). The region only begins to widen at 8B. Transfer is thus trivially perfect in this concentrated regime. Our claim still holds under a compute-time budget.

\begin{table}[!ht]
\centering
\caption{Near-optimal region width (Math, Llama; $\rho=0.5$) when budget is GPU-hour.
A width of $0$ indicates a single near-optimal ratio.}
\label{tab:app_near_optimal_width}
\begin{tabular}{lccc}
\toprule
$\mathcal{R}_\epsilon$
& $\epsilon=2\%$
& $\epsilon=5\%$
& $\epsilon=10\%$ \\
\midrule
1B & $0.00$ & $0.00$ & $0.00$ \\
3B & $0.00$ & $0.00$ & $0.06$ \\
8B & $0.00$ & $0.13$ & $0.25$ \\
\bottomrule
\end{tabular}
\end{table}

\begin{table}[!ht]
\centering
\caption{Cross-scale transfer (Math, Llama; $\rho=0.5$) when budget is GPU-hour.}
\label{tab:app_cross_scale_transfer}
\begin{tabular}{lccc}
\toprule
Transfer
& $\epsilon=0\%$
& $\epsilon=5\%$
& $\epsilon=10\%$ \\
\midrule
1B $\rightarrow$ 3B & $1.00$ & $1.00$ & $1.00$ \\
1B $\rightarrow$ 8B & $1.00$ & $1.00$ & $1.00$ \\
3B $\rightarrow$ 8B & $1.00$ & $1.00$ & $1.00$ \\
\bottomrule
\end{tabular}
\end{table}

\section{Additional Related Work on Mixture Optimization and Scaling Law}
\label{app:related}
\paragraph{Data mixture optimization.}
A parallel literature optimizes the training data mixture \emph{within} a single training stage. Pretraining-stage methods include scaling-law-based mixture extrapolation~\citep{jiang2025adaptive, liu2025regmix, ye2025data, diao2026nemotronclimb, shukorscaling}; group-DRO bi-level domain reweighting~\citep{xie2023doremi, fan2024doge}; online refinement of mixture weights via multi-fidelity Bayesian optimization~\citep{ouyang2025admirebayesopt, yen2026data}, intrinsic-feature signals~\citep{zhu2025toremi}, influence-function attribution~\citep{chang2025automixer}, and gradient statistics~\citep{li2026pike}. 
Post-training mixture methods study data composition or reweighting within individual post-training stages, including SFT data-mixture optimization~\citep{li2025data}, dynamic instruction-tuning mixtures~\citep{zhu2025dynamic}, T\"ulu recipes~\citep{ivison2023camels, lamberttulu3}, DaMo~\citep{shi2025damo}, multi-task preference optimization with AutoMixAlign~\citep{corrado-etal-2025-automixalign}, process-reward-model reweighting with DreamPRM~\citep{cao2025dreamprm}, RLVR data-mixture optimization with MoDoMoDo~\citep{liang2025modomodo}, and recent gradient-based fast mixture solvers~\citep{tan2026fast}.
\citet{bethune2025scaling} is the closest in setting, studying the cross-stages pretraining-finetuning forgetting trade-off via controlled pretraining injection. These methods assume mixture optimization in a single stage, in which weights can be refined as training proceeds, and the same loss surface is shared across mixtures. The SFT--RL allocation problem is fundamentally cross-stage: reallocation requires retraining downstream stages from a different SFT checkpoint; the loss surfaces of SFT and RL differ functionally, ruling out direct application of adaptive or extrapolated point performance along the budget axis. Our work studies a complementary problem: identifying transferable allocation \emph{regions} across model scales at matched budgets.

\paragraph{Scaling laws and cross-scale extrapolation.}
Pretraining scaling laws predict loss from small proxies~\citep{kaplan2020scaling, hoffmann2022training, gadre2025language}, with extensions to loss-to-loss prediction across datasets~\citep{brandfonbrener2025losstoloss} and to downstream task  performance~\citep{ruan2024observational, chen2025scaling}. A related line of work studies whether tuning decisions transfer across scales: \citet{yang2021tuning} shows that under the maximal-update parameterization, optimal hyperparameters can be transferred zero-shot from small to large models in the pretraining setting. 
For the post-training, cross-stage allocation problem we study, we establish transferability empirically rather than through theoretical analysis. 
In post-training, however, scaling behavior is documented to be more fragile: scaling laws can exhibit broken or non-monotonic transitions~\citep{caballero2023broken}, inverse scaling~\citep{mckenzieinverse}, and metric-induced discontinuities~\citep{schaeffer2025why}; reward overoptimization causes performance reversal at high optimization pressure~\citep{gao2023scaling}; and post-training algorithm rankings can flip across scales~\citep{li2026post}. Recent empirical work derives scaling laws specifically for direct alignment~\citep{rafailov2024scaling} and for RLHF data scaling~\citep{shen2026exploring}, but documents bottlenecks that complicate naive extrapolation from small proxies. Our findings sharpen this picture: exact-optimal-ratio transfer can be inconsistent across model scales, but transferring \emph{near-optimal allocation regions} is substantially more reliable, providing a robust mechanism for guiding allocation from small proxy to large target.

\section{Additional Discussion and Clarifications}

\paragraph{Q.1: Why do all experiments use LoRA instead of full-parameter fine-tuning? } ~\\
\textbf{A:} Our analysis sweeps over the Cartesian product of model sizes, allocation ratios, budgets, tasks, algorithms, and model families; on the order of hundreds or even thousands of post-training runs. Full-parameter fine-tuning at this volume can be too expensive, a constraint shared by other recent large-scale post-training studies that use LoRA for the same reason~\citep{raghavendra2025balancing,li2026post}. LoRA has been shown to achieve performance comparable to full fine-tuning across instruction tuning and preference optimization in many settings~\citep{hu2022lora,dettmers2023qlora}. We treat LoRA as a reasonable approximation to full fine-tuning within our experimental scope.

\paragraph{Q.2: How robust are the region width and transferability statistics to training stochasticity?} ~\\
\textbf{A:} Our analysis sweeps post-training runs at the Cartesian product of model sizes, allocation ratios, budgets, tasks, algorithms, and model families: on the order of hundreds or thousands of runs. Multi-seed evaluation across the full sweep would multiply this cost by the number of seeds and would require much more computing resources. To assess whether the qualitative trends we report are robust to seed variance, we conduct a 3-seed validation on math (GSM8K) and summarization for the Llama family (Ablation Study~3 in \cref{app:seed-ablation}); seed variation perturbs both the random subset of training samples drawn from the source pool and the training order within each stage. The seed-averaged performance curves, the near-optimal region width versus tolerance, and the cross-scale transfer hit rates at $\epsilon = 10\%$ all preserve their qualitative trends, confirming that the single-seed findings reported are consistent across seeds.

\paragraph{Q.3: Could the broad near-optimal region be an artifact of the coarse 5-point allocation grid?} ~\\
\textbf{A:} We address this concern in four complementary ways.
\textit{First}, our 5-point grid $\mathcal{G} = \{0.00, 0.25, 0.50, 0.75, 1.00\}$ follows the existing protocol of~\citet{raghavendra2025balancing}, who study the same SFT-vs-preference-finetuning trade-off and report broad allocation trends consistent with our findings. \textit{Second}, we directly validate that the qualitative width and transferability trends are preserved under a denser 9-point grid (adding intermediate ratios $\{0.125, 0.375, 0.625, 0.875\}$) in Ablation Study~1 (\cref{app:dense-grid}); both grids yield the same qualitative scaling and transferability behavior. \textit{Third}, the per-ratio hit-rate heatmaps in our main scaling figures (Row~3 of \cref{fig:main-llama-width} and~\cref{fig:main-qwen-width}) directly visualize the contiguity of the near-optimal region: ratios admitted into $\mathcal{R}_\epsilon$ at a given tolerance are generally neighboring on the grid, indicating genuine smoothness of $\mathcal{P}(N, \cdot, B)$ rather than grid-induced artifacts. \textit{Fourth}, the broad-region phenomenon is observed consistently across three model families (Llama 3, Qwen 2.5, and Qwen 3), four tasks, both off-policy (DPO) and on-policy (GRPO) post-training, and across multiple seeds (Ablation Study~3, \cref{app:seed-ablation}); a grid-induced artifact would not be expected to manifest uniformly across such diverse experimental conditions.

\paragraph{Q.4: Why does the main analysis exclude the low-budget regime ($B < 5\text{k}$)?} ~\\
\textbf{A:} The exclusion is methodologically plausible. \textit{First}, performance curves in the $B < 5\text{k}$ regime can be noisy, with the relative ordering of allocation ratios fluctuating across budgets (per-budget curves in \cref{app:curves}): most prominently for instruction-following on Llama and math on Qwen. Including this regime would conflate the cross-$N$ widening signal with training-instability noise. Similar model-scale transfer works, such as scaling law, routinely exclude warmup-phase~\citep{kaplan2020scaling, hoffmann2022training, caballero2023broken}, and instruction-tuning specifically requires a minimum data threshold~\citep[$\sim$2K examples,]{zhou2023lima}. \textit{Second}, we explicitly extend the analysis to lower budgets in Ablation Study~4 (\cref{app:low-budget}); the qualitative trends persist, with expected attenuation at the noise floor (most visibly for IFEval). At these low budgets, the optimal allocation also shifts toward SFT-heavy ratios~\citep{raghavendra2025balancing}, and direct exhaustive sweeping itself can be cheap. The proxy-transfer story that our framework provides is less load-bearing precisely where the attenuation appears.

\paragraph{Q.5: Why use a relative percentage tolerance rather than an absolute tolerance?} ~\\
\textbf{A:} The relative tolerance, $\mathcal{P}(N, r, B) \geq (1-\epsilon)\mathcal{P}^*(N, B)$, is the practitioner-relevant criterion: deployers can ask ``how much allocation flexibility do I have while retaining $X\%$ of this model's peak?''
An absolute tolerance, $\mathcal{P} \geq \mathcal{P}^* - \delta$, would in principle isolate allocation-sensitivity changes from peak-performance scaling, but defining a meaningful scale-invariant $\delta$ is non-trivial: anchoring $\delta$ to the smallest model collapses to examining relative settings when peak performance is close across scales, and becomes task- and model-dependent (and sometimes ill-defined when the smallest model's performance is at or near trivial baselines). We therefore adopt the relative framing and discuss its consequences. However, the observed widening with scale can be partly driven by peak-performance scaling rather than by changes in allocation sensitivity; a more detailed analysis is provided in \cref{app:abs-tolerance}.

\end{document}